\documentclass[11pt, twoside, a4paper]{book}

\usepackage[pdftitle={The Bees Algorithm for the Vehicle Routing Problem},
pdfauthor={Aish Fenton},
]{hyperref}

\usepackage[ruled]{algorithm2e}
\SetAlgoInsideSkip{smallskip}
\usepackage{amssymb}
\usepackage{amsmath}
\usepackage{amsthm}

\usepackage{graphicx}

\usepackage{float}

\usepackage{pdflscape}
\usepackage{ctable}

\usepackage[margin=40pt,font=small,format=plain,labelfont=bf,textfont=it]{caption}

\newcommand{\CVRP}{\mbox{CVRP}}
\newcommand{\VRP}{\mbox{VRP}}
\newcommand{\MDVRP}{\mbox{MDVRP}}
\newcommand{\VRPTW}{\mbox{VRPTW}}
\newcommand{\PDP}{\mbox{PDP}}
\newcommand{\PDPTW}{\mbox{PDPTW}}
\newcommand{\DARP}{\mbox{DARP}}
\newcommand{\TSP}{\mbox{TSP}}
\newcommand{\MTSP}{\mbox{MTSP}}

\newcommand{\schd}{\mathfrak{S}}

\usepackage{color}

\theoremstyle{definition}

\newcommand{\edge}[1]{(#1)}

\newcommand{\npcomplete}{$\cal NP$-complete}
\newcommand{\nphard}{$\cal NP$-hard}

\newcommand{\BIGO}[1]{\mbox{$O(#1)$}}

\newcommand{\union}{\cup}

\newcommand{\bigunion}{\bigcup}

\newcommand{\intersect}{\cap}

\newcommand{\set}[1]{\{#1\}}
\newcommand{\seq}[1]{[#1]}

\newcommand{\length}[1]{|#1|}

\newcommand{\picscl}[4]{\begin{figure}[htb]\begin{center}\includegraphics[scale=#4]%
	{#1}\caption{#2}\label{#3}\end{center}\end{figure}}

\renewcommand{\implies}{\Rightarrow}

\begin{document}

\pagestyle{empty}
\pagenumbering{roman}


\begin{titlepage}


\vspace*{6em}

\begin{center}
   \LARGE\textbf{The Bees Algorithm for the\\ Vehicle Routing Problem\\}
   \rule{224pt}{0.75pt}\\
   \vspace*{0.75em}
   \rmfamily\large\textit{Aish Fenton}
\end{center}

\hspace{70pt}\includegraphics[scale=0.15, angle=315]{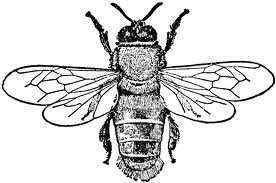}

\vfill
\small\rmfamily
\begin{flushright}
Department of Computer Science\\
University of Auckland \\
Auckland, New Zealand 
\end{flushright}

\end{titlepage}

\texttt{Colophon:}\\
Typeset in \LaTeX\ using typefaces Computer Modern.\\
Title-page image is courtesy of~\cite{beeimage}.


\chapter*{Preface}

This MSc.~thesis has been prepared by Aish Fenton at the University of Auckland, Department of Computer Science. It has been supervised by Dr.~Michael Dinneen. The work undertaken in this thesis has grown out of a research project sponsored by New Zealand Trade and Enterprise (NZTE) for the company vWorkApp Inc.~to research vehicle route optimisation for use within their software product.

\section*{Acknowledgements}

I'd like to thank my partner in crime, Anna Jobsis, for her encouragement, cajoling, threatening, bribing, doing my dishes, proof reading, guilt tripping, grammar policing, comforting\ldots\ doing whatever it took to help me get it done. Also, I'd like to thank my mum and dad for their encouragement and for always making me feel like higher education was within my reach.

Thanks to Steve Taylor and Steve Harding for holding down the fort at work while I disappeared to do this thesis. And likewise thanks to my team, Jono, Rash, Elena, Bob, Marcus, Yuri, and Robin for being on the ball (as always) despite my absence. Thanks to Brendon Petrich and vWorkApp Inc.~for providing me with time off work and being supportive of me undertaking this. 

And lastly, I especially owe Dr.~Michael Dinneen, my supervisor, a big thank you for persevering with me even though he must have doubted that I'd ever finish.

\cleardoublepage

\begin{center}
   \begin{minipage}{300pt}
   \begin{center}
      \textbf{Abstract}
   \end{center}
   In this thesis we present a new algorithm for the Vehicle Routing Problem called the Enhanced Bees Algorithm. It is adapted from a fairly recent algorithm, the Bees Algorithm, which was developed for continuous optimisation problems. We show that the results obtained by the Enhanced Bees Algorithm are competitive with the best meta-heuristics available for the Vehicle Routing Problem---it is able to achieve results that are within 0.5\% of the optimal solution on a commonly used set of test instances. We show that the algorithm has good runtime performance, producing results within 2\% of the optimal solution within 60 seconds, making it suitable for use within real world dispatch scenarios. Additionally, we provide a short history of well known results from the literature along with a detailed description of the foundational methods developed to solve the Vehicle Routing Problem.
   \end{minipage}
\end{center}

\cleardoublepage

\pagenumbering{arabic}
\cleardoublepage
\tableofcontents

\pagestyle{headings}

\chapter{Introduction}

In this thesis we present a new algorithm to solve the Vehicle Routing Problem. The Vehicle Routing Problem describes the problem of assigning and ordering geographically distributed work to a pool of resources. The aim is minimise the travel cost required to complete the work, while meeting any specified constraints, such as a maximum shift duration. Often the context used is that of a fleet of vehicles delivering goods to their customers, although the problem can equally be applied across many different industries and scenarios, and has been applied to non-logistics applications, such as microchip layout. 

Interest in the Vehicle Routing Problem has increased over the last two decades as the cost of transporting and delivering goods has become a key factor of most developed economies. Even a decrease of a few percent on transportation costs can offer a savings of billions for an economy. In the context of New Zealand (our home country) virtually every product grown, made, or used is carried on a truck at least once during its lifetime~\cite{RTFNZ}. The success of New Zealand's export industries are inextricably linked to the reliability and cost effectiveness of our road transport. Moreover, a 1\% growth in national output requires a 1.5\% increase in transport services~\cite{RTFNZ}. 

The Vehicle Routing Problem offers real benefits to transport and logistics companies. Optimisation of the planning and distribution process, such as modelled by the Vehicle Routing Problem, can offer savings of anywhere between 5\% to 20\% of the transportation costs~\cite{TV2001}. Accordingly, the Vehicle Routing Problem has been the focus of intense research since its first formal introduction in the fifties. The Vehicle Routing Problem is one of the most studied of all combinatorial optimisation problems, with hundreds of papers covering it and the family of related problems since its introduction fifty years ago.

The challenge of the Vehicle Routing Problem is that it combines two (or more, in the case of some of its variants) combinatorially hard problems that are themselves known to be \nphard. Its membership in the family of \nphard\ problems makes it very unlikely that an algorithm exists, with reasonable runtime performance, that is able to solve the problem exactly. Therefore heuristic approaches must be developed to solve all but the smallest sized problems.

Many methods have been suggested for solving the Vehicle Routing Problem. In this thesis we develop a new meta-heuristic algorithm that we call the Enhanced Bees Algorithm. We adapt this from another fairly recent algorithm, named the Bees Algorithm, that was developed for solving continuous optimisation problems.

We show that the results obtained by the Enhanced Bees Algorithm are competitive with the best modern meta-heuristics available for the Vehicle Routing Problem. Additionally, the algorithm has good runtime performance, producing results within 2\% of the optimal solution within 60 seconds. This makes the Enhanced Bees Algorithm suitable for use within real world dispatch scenarios where often the dispatch process is dynamic and hence it is impractical for a dispatcher to wait minutes (or hours) for an optimal solution\footnote{The Enhanced Bees Algorithm was developed as part of a New Zealand Trade and Enterprise Research and Development grant for use within the dispatch software product, vWorkApp. Hence more consideration has been given to its runtime performance than is typically afforded in the literature.}.

\section{Content Outline}

We start in Chapter~\ref{chap:background} by providing a short history of the Vehicle Routing Problem, as well as providing background material necessary for understanding the Enhanced Bees Algorithm. In particular, we review the classic methods that have been brought to bear on the Vehicle Routing Problem along with the influential results achieved in the literature. In Chapter~\ref{chap:pd} we provide a formal definition of the Vehicle Routing Problem and briefly describe the variant problems that have been developed in the literature. In Chapter~\ref{chap:algorithm} we provide a detailed description of the Enhanced Bees Algorithm and its operation, along with a review of the objectives that the algorithm is designed to meet, and a description of how the algorithm internally represents the Vehicle Routing Problem. In Chapter~\ref{chap:results} we provide a detailed breakdown of the results obtained by the Enhanced Bees Algorithm. The algorithm is tested against the well known set of test instances due to Christofides, Mingozzi and Toth~\cite{CMT:1981} and is contrasted with some well known results from the literature. Finally, in Chapter~\ref{chap:conclusion} we provide a summary of the results achieved by the Enhanced Bees Algorithm in context of the other methods available for solving the Vehicle Routing Problem from the literature. Additionally, we offer our thoughts on future directions and areas that warrant further research.  


\chapter{Background}
\label{chap:background}

This chapter provides a short history and background material on the Vehicle Routing Problem. In particular we review the solution methods that have been brought to bear on the Vehicle Routing Problem and some of the classic results reported in the literature.

This chapter is laid out as follows. We start in Section~\ref{sec:vo} by informally defining what the Vehicle Routing Problem is and by providing a timeline of the major milestones in its research. We also review a closely related problem, the Traveling Salesman Problem, which is a cornerstone of the Vehicle Routing Problem. We then review in Section~\ref{sec:em} the \emph{Exact~Methods} that have been developed to solve the Vehicle Routing Problem. These are distinguished from the other methods we review in that they provide exact solutions, where the globally best answer is produced. We follow this in Section~\ref{sec:ch} by reviewing the classic \emph{Heuristics} methods that have been developed for the Vehicle Routing Problem. These methods are not guaranteed to find the globally best answer, but rather aim to produce close to optimal solutions using algorithms with fast running times that are able to scale to large problem instances. In Section~\ref{sec:mh} we review \emph{Meta-heuristic} methods that have been adapted for the Vehicle Routing Problem. These methods provide some of the most competitive results available for solving the Vehicle Routing Problem and are considered state-of-the-art currently. Lastly, in Section~\ref{sec:si} we review a modern family of meta-heuristics called \emph{Swarm~Intelligence} that has been inspired by the problem solving abilities exhibited by some groups of animals and natural processes. These last methods have become a popular area of research recently and are starting to produce competitive results to many problems. This thesis uses a Swarm Intelligence method for solving the Vehicle Routing Problem.

\section{Overview}
\label{sec:vo}

The Vehicle Routing Problem (commonly abbreviated to \VRP) describes the problem of assigning and ordering work for a finite number of resources, such that the cost of undertaking that work is minimised. Often the context used is that of a fleet of vehicles delivering goods to a set of customers, although the problem can equally be applied across many different industries and scenarios (including non-logistics scenarios, such as microchip layout). The aim is to split the deliveries between the vehicles and to specify an order in which each vehicle undertakes its work, such that the distance travelled by the vehicles is minimised and any pre-stated constraints are met. In the classic version of the \VRP\ the constraints that must be met are: 

\begin{enumerate}
   \item Each vehicle must start and end its route at the depot.
   \item All goods must be delivered.
   \item The goods can only be dropped off a single time and by a single vehicle.
   \item Each good requires a specified amount of capacity. However, each vehicle has a finite amount of capacity that cannot be exceeded. This adds to the complexity of the problem as it necessarily influences the selection of deliveries assigned to each vehicle.
\end{enumerate}

\picscl{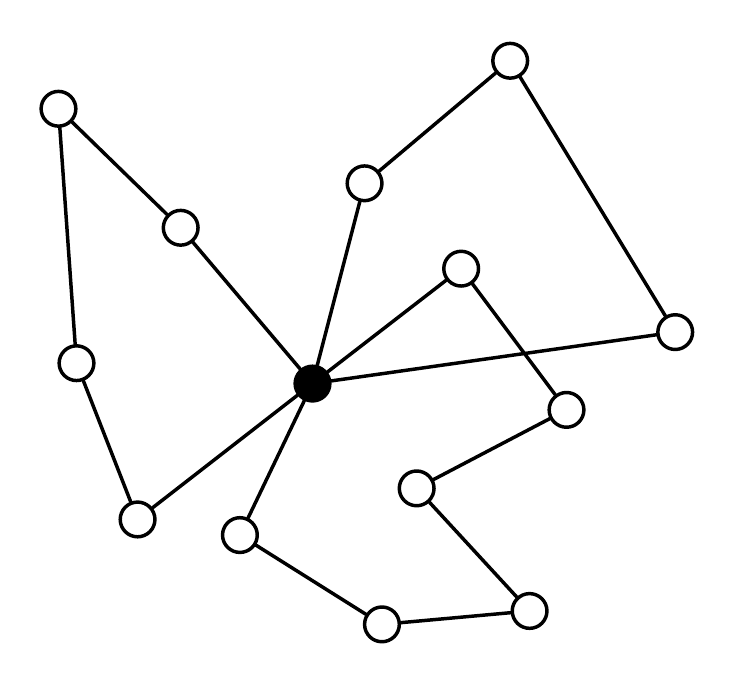}{An example of customers being assigned to three different vehicle routes. The depot is the black dot in the centre.}{fig:vrp_example}{0.55}

More formally, the \VRP\ can be represented as a graph, $(V,E)$. The vertices of the graph, $V$, represent all locations that can be visited; this includes each customer location and the location of the depot. For convenience let $v^d$ denote the vertex that represents the depot. We denote the set of customers as $C = \set{1,2,\ldots,n}$. Next let the set of edges, $E$, correspond to the valid connections between customers and connections to the depot---typically for the \VRP\ all connections are possible. Each edge, $\edge{i,j} \in E$, has a corresponding cost $c_{ij}$. This cost is typically the travel distance between the two locations.

A solution to a given \VRP\ instance can be represented as a family of \emph{routes}, denoted by $\schd$. Each route itself is a sequence of customer visits that are performed by a single vehicle, denoted by $R = \seq{v_1, v_2,\ldots, v_k}$ such that $v_i \in V$, and $v_1, v_k = v^d$. Each customer has a demand $d_i, i \in C$, and $q$ is the maximum demand permissible for any route (i.e.~its maximum capacity). The cost of the solution, and the value we aim to minimise, is given by the following formula:

\[
   \sum_{R \in \schd} \sum_{v_i \in R} c_{v_i, v_{i+1}}
\]

We can now formalise the \VRP\ constraints as follows:

\begin{align}
   & \bigunion_{R \in \schd} = V \label{eq:vf1}\\
   & (R_i - v^d) \intersect R_j = \emptyset  && \forall R_i, R_j \in \schd \label{eq:vf2}\\
   & v_i = v_j                               && \forall R_i \in \schd, \lnot \exists v_i, v_j \in (R_i - v^d) \label{eq:vf3}\\
   & v_0, v_k \in R_i = v^d                  && \forall R_i \in \schd \label{eq:vf4}\\
   & \sum_{v \in R_i} d_v < q                && \forall R_i \in \schd \label{eq:vf5}
\end{align}

Equation~\eqref{eq:vf1} specifies that all customers are included in at least one route. Equations~\eqref{eq:vf2} and~\eqref{eq:vf3} ensure that each customer is only visited once, across all routes. Equation~\eqref{eq:vf4} ensures that each route starts and ends at the depot. Lastly, Equation~\eqref{eq:vf5} ensures that each route doesn't exceed its capacity.

This version of the problem has come to be known as the Capacitated Vehicle Routing Problem (often appreciated to \CVRP\ in the literature). See Chapter~\ref{chap:pd} for an alternative formation, which states the problem as an Integer Linear Programming problem, as is more standard in the \VRP\ literature\footnote{We believe that the formation provided in this chapter is simpler and more precise for understanding the algorithmic methods described in this chapter. However, we do provide a more standard formation in Chapter~\ref{chap:pd}.}.

\VRP\ was first formally introduced in 1959 by Dantzig and Ramser in their paper, the \emph{Truck~Scheduling~Problem}~\cite{Dantzig:1959}. The \VRP\ has remained an important problem in logistics and transport, and is one of the most studied of all combinatorial optimisation problems. Hundreds of papers have been written on it over the intervening fifty years. From the large number of implementations in use today it is clear that the \VRP\ has real benefits to offer transport and logistics companies. Anywhere from 5\% to 20\% savings have been reported where a vehicle routing procedure has been implemented~\cite{TV2001}.

From the \VRP\ comes a family of related problems. These problems model other constraints that are encountered in real world applications of the \VRP. Classic problems include: \VRP\ with Time Windows (\VRPTW), which introduces a time window constraint against each customer that the vehicle must arrive within; \VRP\ with Multiple Depots (\MDVRP), where the vehicles are dispatched from multiple starting points; and the Pickup and Delivery Problem (\PDP), where goods are both picked up and delivered during the course of the route (such as a courier would do).

\subsection{\TSP\ Introduction and History}
\label{sec:tiah}

The \VRP\ is a combination of two problems that are combinatorial hard in themselves: the Traveling Salesman Problem (more precisely the Multiple Traveling Salesman Problem), and the Bin Packing Problem.

The Traveling Salesman Problem (\TSP) can informally be defined as follows. Given $n$ points on a map, provide a route through each of the $n$ points such that each point is only used once and the total distance travelled is minimised. The problem's name, the \emph{Traveling~Salesman}, comes from the classic real world example of the problem. A salesman is sent on a trip to visit $n$ cities. They must select the order in which to visit the cities, such that they travel the least amount of distance.

Although the problem sounds like it might be easily solvable, it is in fact \nphard. The best known exact algorithms for solving the \TSP\ still require a running time of $\BIGO{2^n}$. Karp's famous paper, \emph{Reducibility~Among~Combinatorial~Problems}~\cite{Kar72}, in 1972 showed that the Hamiltonian Circuit problem is \npcomplete. This implied the NP-hardness of \TSP, and thus supplied the mathematical explanation for the apparent difficulty of finding optimal traveling salesman tours. 

\TSP\ has a history reaching back many years. It is itself related to another classic graph theory problem, the Hamiltonian circuit. Hamiltonian circuits have been studied since 1856 by both Hamilton~\cite{Hamilton:1856} and Kirkman~\cite{Kirkman:1856}. Whereas the \TSP\ has been informally discussed for many years~\cite{Schrijver}, it didn't become actively studied until after 1928, where Menger, Whitney, Flood and Robinson produced much of the early results in the field. Robinson's RAND report~\cite{Robinson:1949} is probably the first article to call the problem by the name it has since become known as, the Traveling Salesman Problem. 

\clearpage

From Robinson's RAND report:
\begin{quote}
\itshape
   The purpose of this note is to give a method for solving a problem related to the traveling salesman problem. One formulation is to find the shortest route for a salesman starting from Washington, visiting all the state capitals and then returning to Washington. More generally, to find the shortest closed curve containing $n$ given points in the plane.
\end{quote}

\picscl{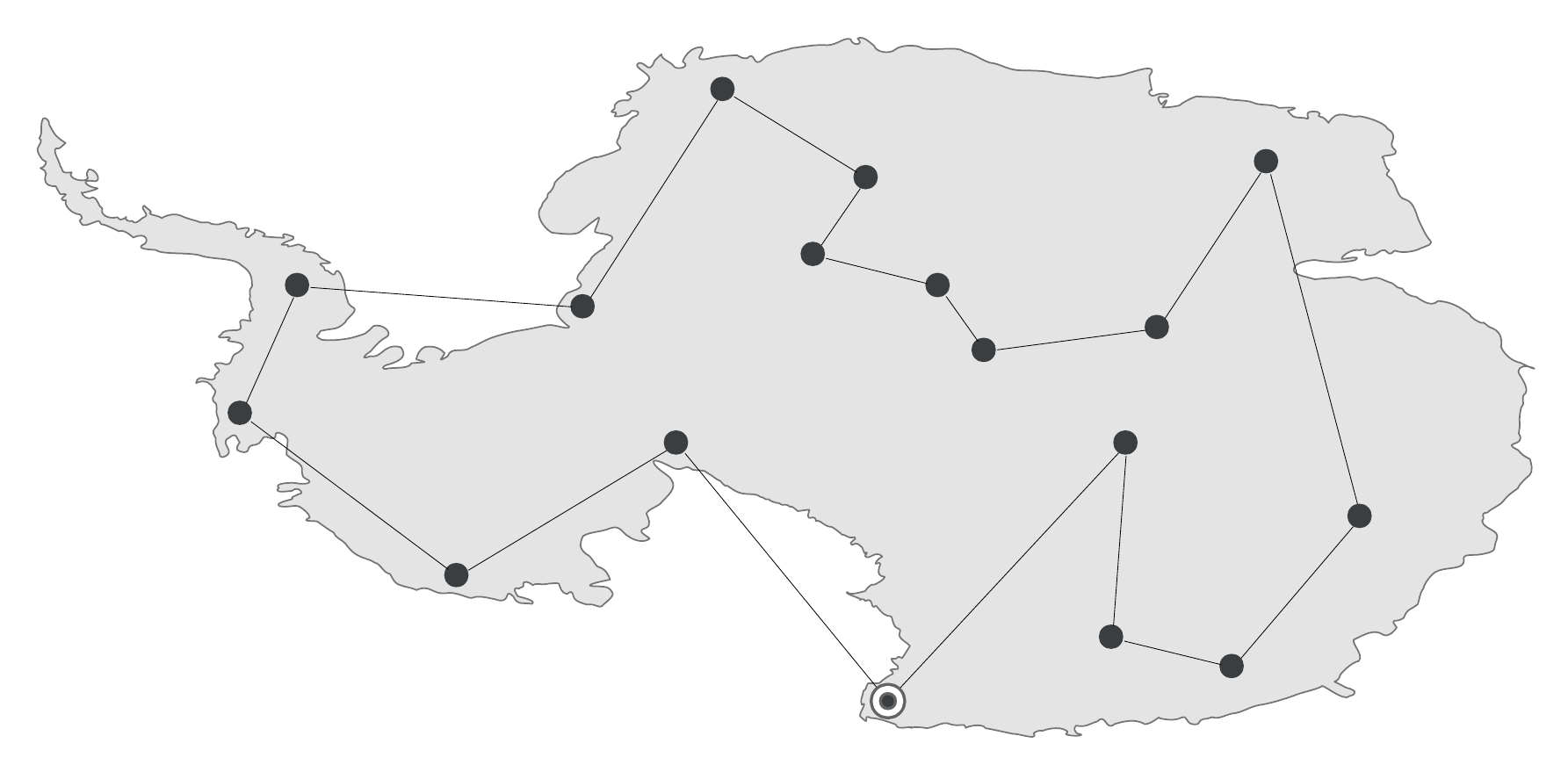}{Shown is an example of a 16 city \TSP\ tour. This tour is one of 20,922,789,888,000 possible tours for these 16 cities.}{fig:tsp}{0.60}

An early result was provided by Dantzig, Fulkerson, and  Johnson~\cite{Dantzig:1954}. Their paper gave an exact method for solving a 49 city problem, a large number of cities for the time. Their algorithm used the cutting plane method to provide an exact solution. This approach has been the inspiration for many subsequent approaches, and is still the bedrock of algorithms that attempt to provide an exact solution.  

A generalisation of the \TSP\ is Multiple Traveling Salesman Problem (\MTSP), where multiple tours are constructed (i.e.~multiple salesman can be used to visit the cities). The pure \MTSP\ can trivially be turned into a \TSP\ by constructing a graph $G$ with $n - 1$ additional copies of the starting vertex and by forbidding travel directly between the $n$ starting vertices. However, the pure formulation of \MTSP\ places no additional constraints on how the routes are constructed. Real life applications of the \MTSP\ typically require additional constraints, such as limiting the size or duration of each route (i.e.~one salesman shouldn't be working a 12 hour shift, while another has no work).

\MTSP\ leads us naturally into the family of problems given by the \VRP. \VRP, and its family of related problems, can be understood as being a generalisation of \MTSP\ that incorporates additional constraints. Some of these constraints, such as capacity limits, introduce additional dimensions to the problem that are in themselves hard combinatorial problems.

\section{Exact Methods}
\label{sec:em}

The first efforts at providing a solution to the \VRP\ were concerned with exact methods. These started by sharing many of the techniques brought to bear on \TSP. We follow Laporte and Nobert's survey~\cite{LANO:87} and classify exact methods for the \VRP\ into three families: Direct Tree Search methods, Dynamic Programming, and Integer Linear Programming.

The first classic Direct Tree Search results are due to Christolds and Ellison. Their 1969 paper provided the first branch and bound algorithm for exactly solving the \VRP~\cite{CE:1969}. Unfortunately its time and memory requirements were large enough that it was only able to solve problems of up to 13 customers. This result was later improved upon by Christolds in 1976 by using a different branch model. This improvement allowed him to solve for up to 31 customers. 

Christofides, Mingozzi, and Toth~\cite{CMT:1981}, provide a lower bound method that is sufficiently quick (in terms of runtime performance) to be used as a lower bound for excluding nodes from the search tree. Using this lower bound they were able to provide solutions for problems containing up to 25 customers. Laporte, Mercure and Nobert~\cite{LMN:1986} used \MTSP\ as a relaxation of the \VRP\ within a branch and bound framework to provide solutions for more realistically sized problems, containing up to 250 customers.  

A Dynamic Programming approach was first applied to the \VRP\ by Eilon, Watson-Gandy and Christofides~\cite{EWC:1971}. Their approach allowed them to solve exactly for problems of 10 to 25 customers. Since then, Christofides has made improvements to this algorithm to solve exactly for problems up to fifty customers.

A Set Partitioning method was given by Balinski, and Quandt in 1964~\cite{balinski:64} to produce exact \VRP\ solutions. However, the problem sets they used were very small, only containing between 5 to 15 customers; and even then they were not able to produce solutions for some of the problems. However, taking their approach as a starting point, many authors have been able to produce more powerful methods. Rao and Zionts~\cite{RZ:1968}, Foster and Ryan~\cite{FR:1976}, and Desrochers, Desrosiers and Solomon~\cite{DMDJSM:1992} have all extended the basic set partitioning algorithm using the Column Generation method from Integer Programming. These later papers have produced some of the best exact results. 

Notwithstanding the preceding discussion, exact methods have been of more use in advancing the theoretical understanding of the \VRP\ than they have been in providing solutions to real life routing problems. This can mostly be attributed to the fact that real life \VRP\ instances often involve at least tens of customers (and often hundreds), and involve richer constraints than are modelled in the classic \VRP.

\section{Classic Heuristics}
\label{sec:ch}

In this section we review the classic heuristic methods that have been developed for the \VRP. These methods are not guaranteed to find the globally best answer, but rather aim to produce close to optimal solutions using algorithms with fast running times that are able to scale to large problem instances. Classic heuristics for the \VRP\ can be classified into three families: constructive heuristics; two-phase heuristics, which can again be divided into two subfamilies, cluster first and then route, and route first and then cluster; and finally improvement methods.

\subsection{Constructive Heuristics}
\label{subsec:conheu}

We start by looking at the Constructive Heuristics. Constructive heuristics build a solution from the ground up. They typically provide a recipe for building each route, such that the total cost of all routes is minimised.

A trivial but intuitive constructive heuristic is the \emph{Nearest~Neighbour} method. In this method routes are built up sequentially. At each step the customer nearest to the last routed customer is chosen. This continues until the route reaches its maximum capacity, at which point a new route is started. In practice the Nearest Neighbour algorithm tends to provide poor results and is rarely used. 
 
\picscl{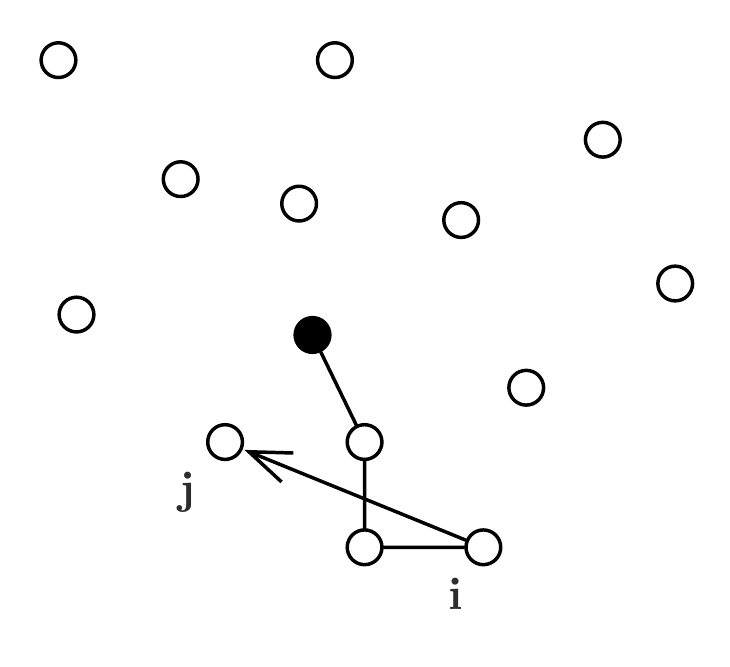}{Shown is an example of the Nearest Neighbour method being applied. A partially constructed route selects customer $j$ to add, as its closest to the last added customer, $i$.}{fig:nn}{0.55}

An early and influential result was given by Clarke and Wright in their 1964 paper~\cite{clark:1964}. In their paper they present a heuristic extending Dantzig and Ramser's earlier work, which has since become known as the \emph{Clarke~Wright~Savings} heuristic. The heuristic is based on the simple premise of iteratively combining routes in order of those pairs that provide the largest saving. 

\picscl{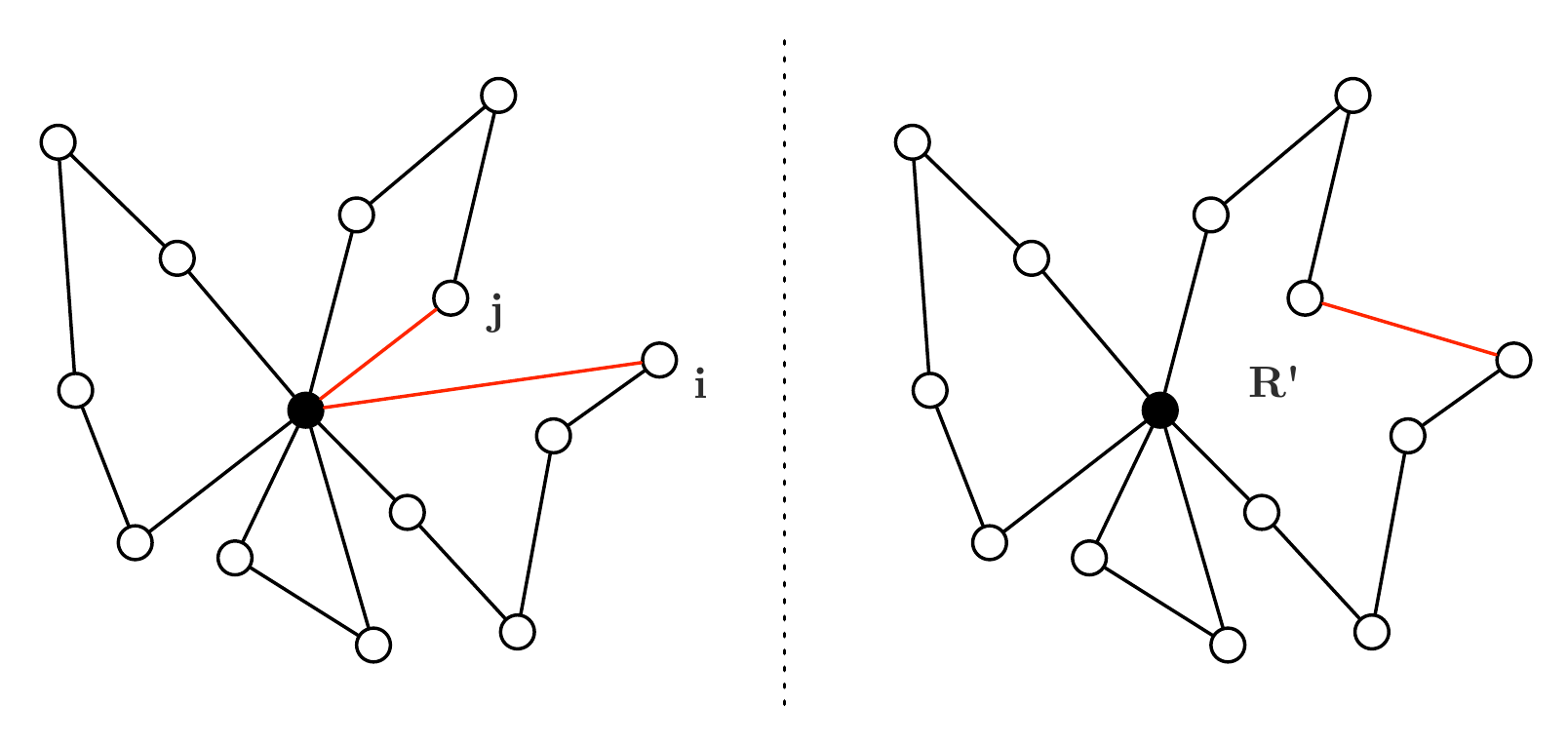}{Clark Wright Savings Algorithm. Customers $i,j$ are selected as candidates to merge. The merge results in a new route $R'$.}{fig:cw_savings}{0.55}

The algorithm works as follows:

\begin{algorithm}[H]
   \caption{Clark Write Savings Algorithm}
   initialiseRoutes()\\
   $M$ = savingsMatrix($V$)\\
   $L$ = sortBySavings($SM$)\\
   \For{$l_{ij} \leftarrow L$}{
      $R^i, R^j$ = findRoutes($l_{ij}$)\\
      \If{feasibleMerge($R^i, R^j$)}{
         combineRoute($R^i, R^j$)
      }
   }
\end{algorithm}

The algorithm starts by initialising a candidate solution. For this it creates a route $R = \seq{v^d, v_i, v^d}$ for all $v \in V$. It then calculates a matrix $M$ that contains the savings $s_{ij} = c_{i0} + c_{j0} - c_{ij}$ for all edges $\edge{i,j} \in E$. It then produces a list, $L$, that enumerates each cell $i,j$ of the matrix in descending order of the savings. For each entry in the list, $l_{ij} \in L$, it selects the two routes, $R^i, R^j$, that contain customers $i,j \in V$ and tests to see if the two routes can be merged. A merge is permissible if and only if:

\begin{enumerate}
   \item $R^i \neq R^j$.
   \item $i,j$ are the first or last vertices (excluding the depot $v^d$) of their respective routes.
   \item The combined demand of the two routes doesn't exceed the maximum allowed, $q$.
\end{enumerate}

The heuristic comes in two flavours, sequential and parallel. The sequential version adds the additional constraint that only one route can be constructed at a time. In this case one of the two routes considered, $R^i, R^j$, must be the route under construction. If neither of the routes are the route under construction then the list item is ignored and processing continues down the list. If the merge is permissible then we merge routes $R^i, R^j$ such that $R' = \seq{v_0,\ldots,i,j,\ldots,v_k}$. In the parallel version, once the entire list of savings has been enumerated then the resulting solution is returned as the answer. In the sequential version the \emph{for~loop} is repeated until no feasible merges remain.

The Clark Write Savings heuristic has been used to solve problems of up to 1000 customers with results often within 10\% of the optimal solution using only a 180 seconds of runtime~\cite{TV2001}. The parallel version of the Clark Write Savings Algorithm outperforms the sequential version in most cases~\cite{Laporte:1999} and is typically the one employed.

The heuristic has proven to be surprisingly adaptable and has been extended to deal with more specialised vehicle routing problems where additional objectives and constraints must be factored in. Its flexibility is a result of its algebraic treatment of the problem~\cite{Laporte:1999}. Unlike many other \VRP\ heuristics that exploit the problem's spatial properties (such as many of the two-phase heuristics, see Section~\ref{subsec:tph}), the savings formula can easily be adapted to take into consideration other objectives. An example of this is Solomon's equally ubiquitous algorithm~\cite{Solomon:1987} which extends the Clark Wright Savings algorithm to cater for time constraints. 

This classic algorithm has been extended by Gaskell~\cite{Gaskell:1967}, Yellow~\cite{Yellow:1970} and Paessens~\cite{Paessens:1988}, who have suggested alternatives to the savings formulas used by Clarke and Wright. These approaches typically introduce additional parameters to guide the algorithm towards selecting routes with geometric properties that are likely to produce better combinations. Altinkemer and Gavish provide an interesting variation on the basic savings heuristic~\cite{AG:1991}. They use a matching algorithm to combine multiple routes in each step. To do this they construct a graph such that each vertex represents a route, each edge represents a feasible saving, and the edges' weights represent the savings that can be realised by the merge of the two routes. The algorithm proceeds by solving a maximum cost weighted matching of the graph.

\subsection{Two-phase Heuristics}
\label{subsec:tph}

We next look at two-phase heuristics. We start by looking at the cluster first, route second subfamily. One of the foundational algorithms for this method is given to us by Gillett and Miller who provided a new approach called the Sweep Algorithm in their 1974 paper~\cite{GM:1974}. This popularised the two-phase approach, although a similar method was suggested earlier by Wren in his 1971 book, and subsequently in Wren and Holliday's 1972 paper~\cite{WH:1972}. In this approach, an initial clustering phase is used to cluster the customers into a base set of routes. From here the routes are treated as separate \TSP\ instances and optimised accordingly. The two-phase approach typically doesn't prescribe a method for how the \TSP\ is solved and assumes that already developed \TSP\ methods can be used. The classic \emph{Sweep} algorithm uses a simple geometric method to cluster the customers. Routes are built by sweeping a ray, centered at the depot, clockwise around the space enclosing the problem's locations. The Sweep method is surprisingly effective and has been shown to solve several benchmark \VRP\ problems to within 2\% to 9\% of the best known solutions~\cite{TV2001}.

\picscl{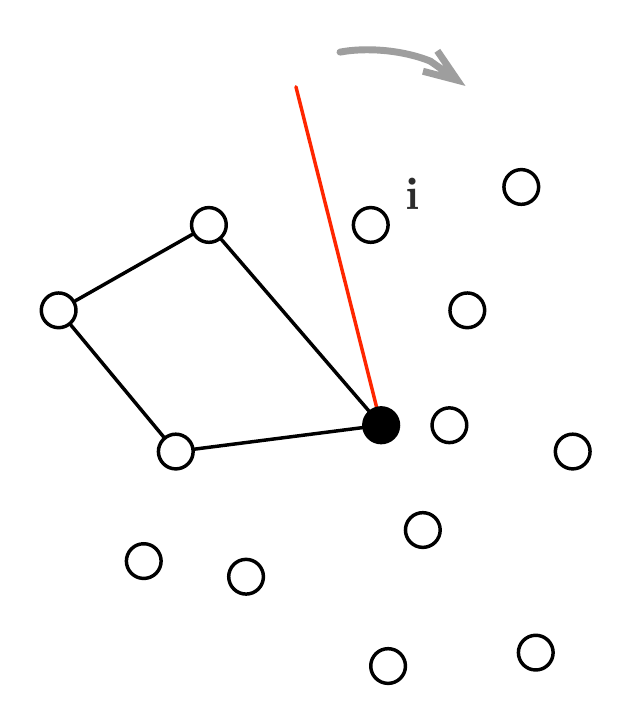}{This diagram shows an example of the Sweep process being run. The ray is swept clockwise around the geographic area. In this example one route has already been formed, and a second is about to start at customer $i$.}{fig:sweep}{0.55}

Fisher and Jaikumars's 1981 paper~\cite{FJ:1981} builds upon the two-phase approach by providing a more sophisticated clustering method. They solve a General Assignment Problem to form the clusters instead. A limitation of their method is that the amount of vehicle routes must be fixed up front. Their method often produces results that are 1\% to 2\% better than results produced by the classic Sweep algorithm~\cite{TV2001}. 

Christofides, Mingozzi, and Toth expanded upon this approach in~\cite{CMT:1981} and proposed a method that uses a truncated branch and bound technique (similar to Christofides's Exact method). At each step it builds a collection of candidate routes for a particular customer, $i$. It then evaluates each route by solving it as a \TSP, from which it then selects the shortest \TSP\ as the route.

The Petal algorithm is a natural extension to the Sweep algorithm. It was first proposed by Balinski and Quandt~\cite{balinski:64} and then extended by Foster and Ryan~\cite{FR:1976}. The basic process is to produce a collection of overlapping candidate routes (called petals) and then to solve a set partitioning problem to produce a feasible solution. As with other two-phase approaches it is assumed that the order of the customers within each route is solved using an existing \TSP\ heuristic. The petal method has produced competitive results for small solutions, but quickly becomes impractical where the set of candidate routes that must be considered is large.  

Lastly, there are route first, cluster second methods. The basic premise of these techniques are to first construct a `grand' \TSP\ tour such that all customers are visited. The second phase is then concerned with splitting this tour into feasible routes. Route first, cluster second methods are generally thought to be less competitive than other methods~\cite{Laporte:1999}, although interestingly, Haimovich and Rinnooy Kan have shown that if all customers have unit demand then a simple shortest path algorithm (which can be solved in polynomial time) can be used to produce a solution from a \TSP\ tour that is asymptotically optimal~\cite{HK:1985}.

\subsection{Iterative Improvement Heuristics}

Iterative Improvement methods follow an approach where an initial candidate solution is iteratively improved by applying an operation that improves the candidate solution, typically in a small way, many thousands of times. The operations employed are typically simple and only change a small part of the candidate solution, such as the position of a single customer or edge within the solution. The set of solutions that are obtainable from the current candidate solution, $S$, by applying an operator $Op$ is known as $S$'s neighbourhood. Typically, with Iterative Improvement heuristics, a new solution $S'$ is selected by exhaustively searching the entire neighbourhood of $S$ for the best improvement possible. If no improvement can be found then the heuristic terminates. The initial candidate solution (i.e.~the starting point of the algorithm) can be randomly selected or can be produced using another heuristic. Constructive Heuristics are typically used for initially seeding an improvement heuristic, see Section~\ref{subsec:conheu} for more information on these.
 
Probably one of the best known improvement operators is 2-Opt. The 2-Opt operator takes two edges $\edge{i,j}, \edge{k,l} \in T$, where $T$ are the edges traversed by a particular route $R = \seq{v_1,\ldots, v_i, v_j,\ldots,v_k, v_l,\ldots, v_n}$, and removes these from the candidate solution. This splits the route into two disconnected components, $D_1 = \seq{v_j,\ldots, v_k}, D_2 = \seq{v_1,\ldots, v_i,v_l,\ldots, v_n}$. A new candidate solution is produced by reconnecting $D_1$ to $D_2$ using the same vertices $i,j,l,k$ but with alternate edges, such that $\edge{i,j}, \edge{j,l} \in T$.

\picscl{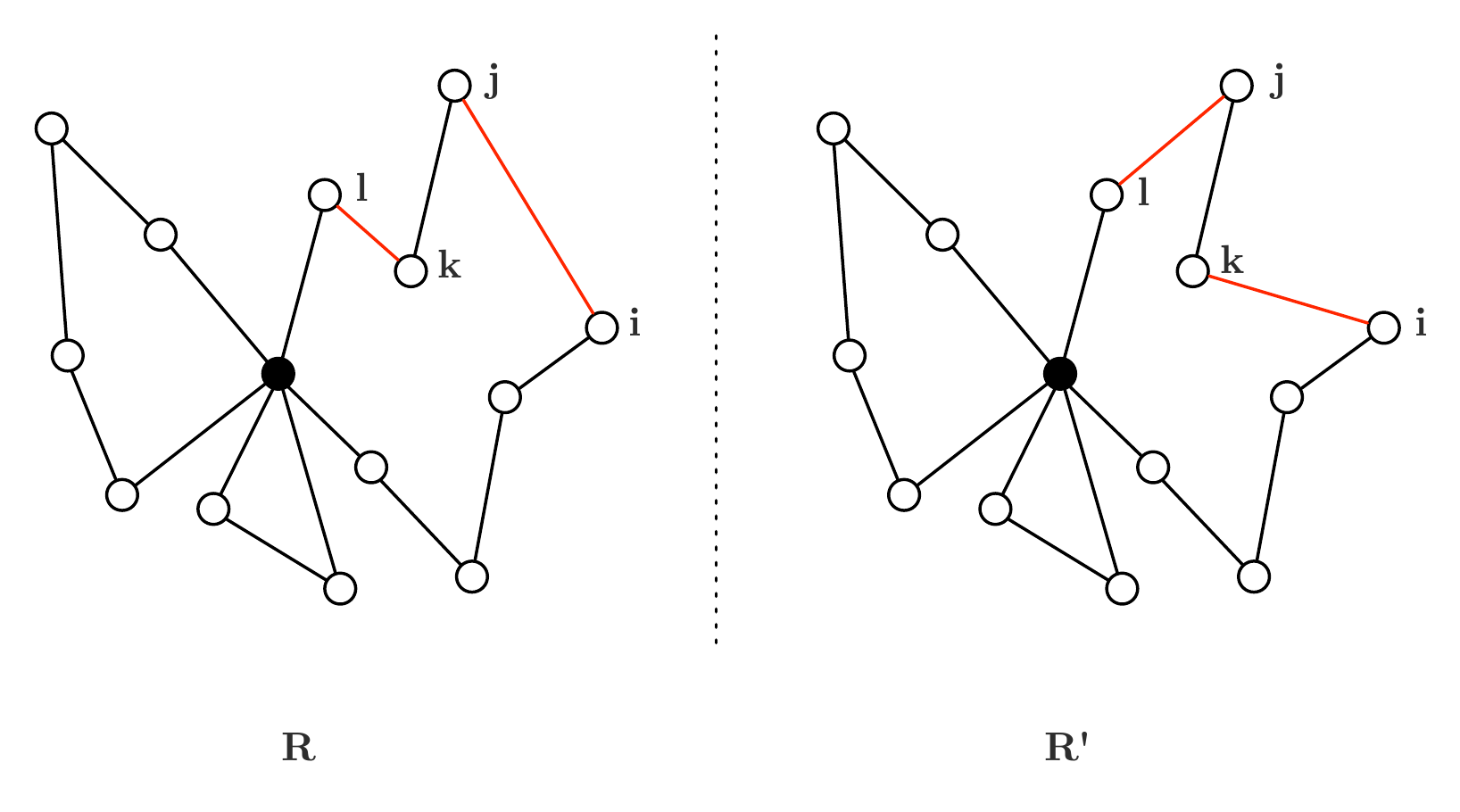}{This diagram shows 2-Opt being applied to a candidate solution $R$ and producing a new solution $R'$. In this example edges $\edge{i,j}, \edge{k,l}$ are exchanged with edges $\edge{i,k}, \edge{j,l}$.}{fig:2opt}{0.55}

The rationale behind 2-Opt is that, due to the triangle inequality, edges that cross themselves are unlikely to be optimal. 2-Opt aims to detangle a route.

There are a number of other operations suggested in the literature. Christofides and Eilon give one of the earliest iterative improvement methods in their paper~\cite{CE:1969}. In the paper they make a simple change to 2-Opt to increase the amount of edges removed from two to three---the operation fittingly being called 3-Opt. They found that their heuristic produced superior results than 2-Opt.

In general, operations such as 3-Opt, that remove edges and then search for a more optimal recombination of components take \BIGO{n^y} where $y$ is the number of edges removed. A profitable strain of research has focused on producing operations that reduce the amount of recombinations that must be searched. Or presents an operation that has since come to be known as Or-Opt~\cite{Or:1976}. Or-Opt is a restricted 3-Opt. It searches for a relocation of all sets of 3 consecutive vertices (which Or calls chains), such that an improvement is made. If an improvement cannot be made then it tries again with chains of 2 consecutive vertices, and so on. Or-Opt has been shown to produce similar results to that of 3-Opt, but with a running time of \BIGO{n^2}. More recently Renaud, Boctor, and Laptorte~\cite{RBL:1996} have presented a restricted version of 4-Opt, called 4-Opt*, that operates in a similar vein to Or-Opt. 4-Opt* has a running time of \BIGO{wn^2} where $w$ denotes the number of edges spanned by 4-Opt* when building a chain. 

Iterative improvement heuristics are often used in combination with other heuristics. In this case they are run on the candidate solution after the initial heuristic has completed. However, if used in this way there is often a fine balance between producing an operation that improves a solution, and one that is sufficiently destructive enough to escape a local minimum. Interest in Iterative Improvement heuristics has grown as the operations developed for them, such as Or-Opt, are directly applicable to more modern heuristics, such as the family known as meta-heuristics presented in the next section. 

\section{Meta-heuristics}
\label{sec:mh}

Meta-heuristics are a broad collection of methods that make few or no assumptions about the type of problem being solved. They provide a framework that allows for individual problems to be modelled and `plugged in' to the meta-heuristic. Typically,  meta-heuristics take an approach where a candidate solution (or solutions) is initially produced and then is iteratively refined towards the optimal solution. Intuitively meta-heuristics can be thought of searching a problem's search space. Each iteration searches the neighbourhood of the current candidate solution(s) looking for new candidate solutions that move closer to the global optimum.

\picscl{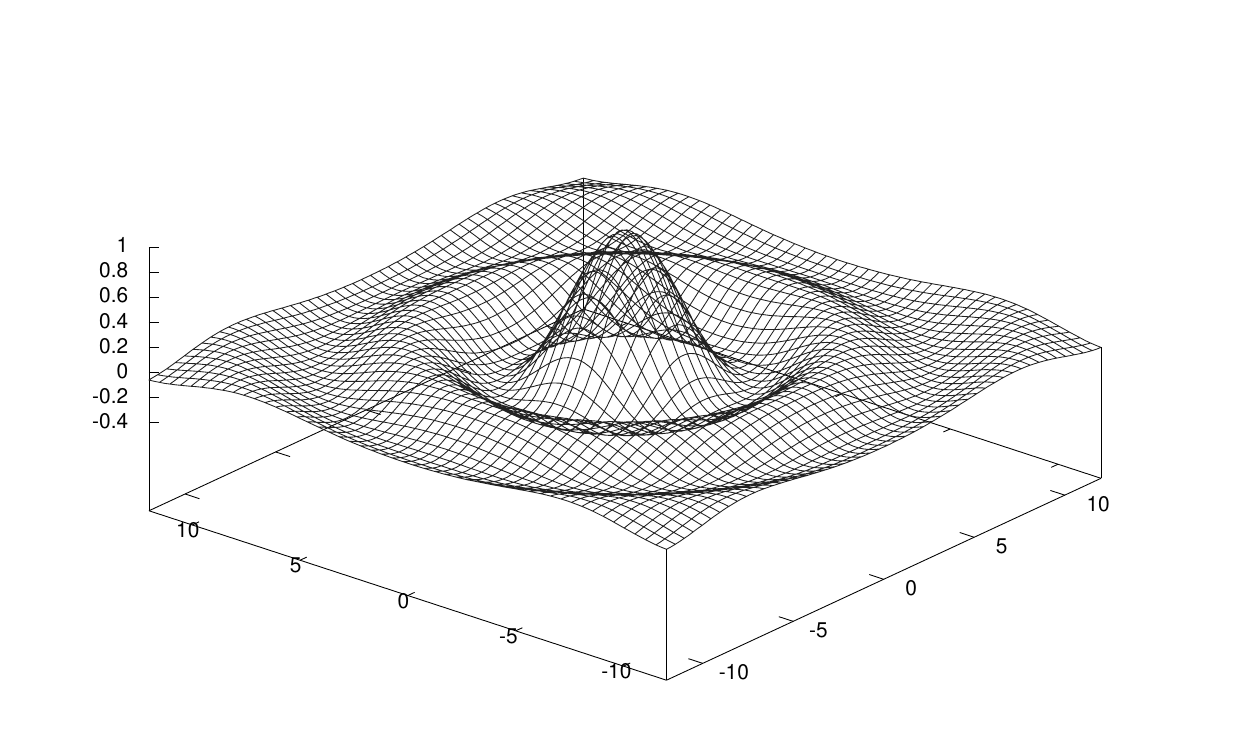}{This diagram shows an example of a search space that a meta-heuristic moves through. In this example the peak at the centre of the figure is the globally best answer, but there are also hills and valleys that the meta-heuristic may become caught in. These are called local minima and maxima.}{fig:gss}{0.66}

A limitation of meta-heuristics is that they are not guaranteed to find an optimal solution (or even a good one!). Moreover, the theoretical underpinnings of what makes one meta-heuristic more effective than another are still poorly understood. Meta-heuristics within the literature tend to be tuned for specific problems and then validated empirically.

There have been a number of meta-heuristics produced for the \VRP\ in recent years and many of the most competitive results produced in the last ten years are due to them. We next review some of the more well known meta-heuristic results for the \VRP.

\subsection{Simulated Annealing}
\label{subsec:simulatedannealing}

Simulated Annealing is inspired by the annealing process used in metallurgy. The algorithm starts with a candidate solution (which can be randomly selected) and then moves to nearby solutions with a probability dependent on the quality of the solution and a global parameter $T$, which is reduced over the course of the algorithm. In classic implementations the following formula is used to control the probability of a move: 

\[
e^{-\frac{f(s')-f(s)}{T}}
\]

Where $f(s)$ and $f(s')$ represent the solution quality of the current solution, and the new solution respectively. By analogy with the metallurgy process, $T$ represents the current temperature of the solution. Initially $T$ is set to a high value. This lets the algorithm free itself from any local optima that it may be caught in. It is then cooled over the course of the algorithm forcing the search to converge on a solution. 

One of the first Simulated Annealing results for the \VRP\ was given by Robuste, Daganzo and Souleyrette~\cite{RDS:1990}. They define the search neighbourhood as being all solutions that can be obtained from the current solution by applying one of two operations: relocating part of a route to another position within the same route, or exchanging customers between routes. They tested their solution on some large real world instances of up to 500 customers. They reported some success with their approach, but as their test cases were unique, no direct comparison is possible. 

Osman has given the best known Simulated Annealing results for the \VRP~\cite{Osman:1993}. His algorithm expands upon many areas of the basic Simulated Annealing approach. The method starts by using the Clark and Wright algorithm to produce an initial position. It defines its neighbourhood as being all candidate solutions that can be reached by applying an operator he names the $\lambda$-interchange operation.

$\lambda$-interchange works by selecting two sequences (i.e.~chains) of customers $C_p, C_q$ from two routes, $R_p$ and $R_q$, such that $\length{C_p}, \length{C_q} < \lambda$ (note that the chains are not necessarily of the same length). The customers within each chain are then exchanged with each other in turn, until an exchange produces an infeasible solution. As the neighbourhood produced by $\lambda$-interchange is typically quite large, Osman restricts $\lambda$ to being less than 2 and suggests that the first move that provides an improvement is used rather than exhaustively searching the entire neighbourhood.

\picscl{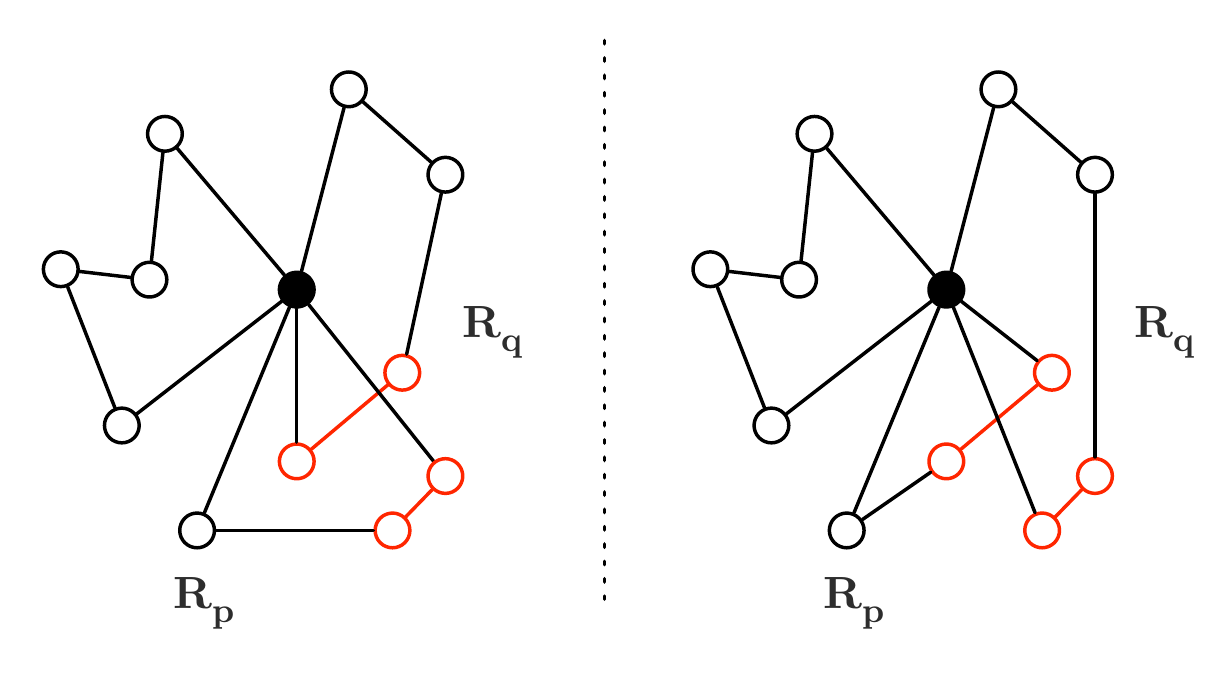}{This diagram shows an example of $\lambda$-interchange being applied to a candidate solution. Two sequences of customers $C_p$ and $C_q$ are selected from routes $R_p$, $R_q$ respectively. Customers from $C_p$ are then swapped with $C_q$ where feasible.}{fig:lambdaopt}{0.66}

Osman also uses a sophisticated cooling schedule. His main change being that the temperature is cooled only while improvements are found. If no improvement is found he then resets the temperature using $T_i = max(\frac{T_r}{2}, T_b)$, where $T_r$ is the reset temperature, and $T_b$ is temperature of the best solution found so far. 

Although Simulated Annealing has produced some good results, and in many cases outperforms classic heuristics (compare~\cite{Laporte:1999} with~\cite{GLP:1999}), it is not competitive with the Tabu Search methods discussed in Section~\ref{sec:ts}.

\subsection{Genetic Algorithms}

Genetic Algorithms were first proposed in~\cite{Holland:1975}. They have since been applied to many problem domains and are particularly well suited to applications that must work across a number of different domains. In fact they were the first evolutionary-inspired algorithm to be applied to combinatorial problems~\cite{Potvin:2009}. The basic operation of a Genetic Algorithm is as follows:

\begin{algorithm}[H]
   \caption{Simple Genetic Algorithm}
   Generate the initial population\\
   \While{termination condition not met}{
      Evaluate the fitness of each individual\\
      Select the fittest pairs\\
      Mate pairs and produce next generation\\
      Mutate (optional)\\
   }
\end{algorithm}

In a classic Genetic Algorithm each candidate solution is encoded as a binary string (i.e.~chromosome). Each individual (i.e.~candidate solution) is initially created randomly and used to seed the population. A technique often employed in the literature is to initially `bootstrap' the population by making use of another heuristic to produce the initial population. However, special care must be taken with this approach to ensure that diversity is maintained across the population, as you risk premature convergence by not introducing enough diversity in the initial population. 

Next, the fittest individuals are selected from the population and are mated in order to produce the next generation. The mating process uses a special operator called a \emph{crossover} operator that takes two parents and produces offspring from these by combining parts of each parent. Optionally, a mutation operation is also applied, that introduces a change that doesn't exist in either parent. The classic crossover operation takes two individuals encoded as binary strings and splits these at one or two points along the length of the string. The strings are then recombined to form a new binary string, which in turn encodes a new candidate solution. The entire process is continued until a termination condition is met (often a predetermined running time), or until the population has converged on a single solution.

Special consideration needs to be given to how problems are encoded and to how the crossover and mutation operators work when using Genetic Algorithms to solve discrete optimisation problems, such as the \VRP. For example, the classic crossover operation, which works on binary strings, would not work well on a \TSP\ tour. When two components of two tours are combined in this way they are likely to contain duplicates. Therefore, it is more common for the \VRP\ (and the \TSP) to use a direct representation and to use specially designed crossover operators. In this instance the \VRP\ is represented as a set of sequences, each holding an ordered list of customers. The crossover operators are then designed so that they take into consideration the constraints of the \VRP.

Two crossover operators commonly used with combinatorial problems are the Order Crossover ($OX$) and the Edge Assembly Crossover ($EAX$). $OX$~\cite{OSH:1987} operates by selecting two cut points within each route. The substring between the two cut points is copied from the second parent directly into the offspring. Likewise, the string outside the cut points is copied from the first parent into the offspring, but with any duplicates removed. This potentially leaves a partial solution, where not all customers have been routed. The partial solutions is then repaired by inserting any unrouted customers into the child in the same order that they appeared in the second parent.

\picscl{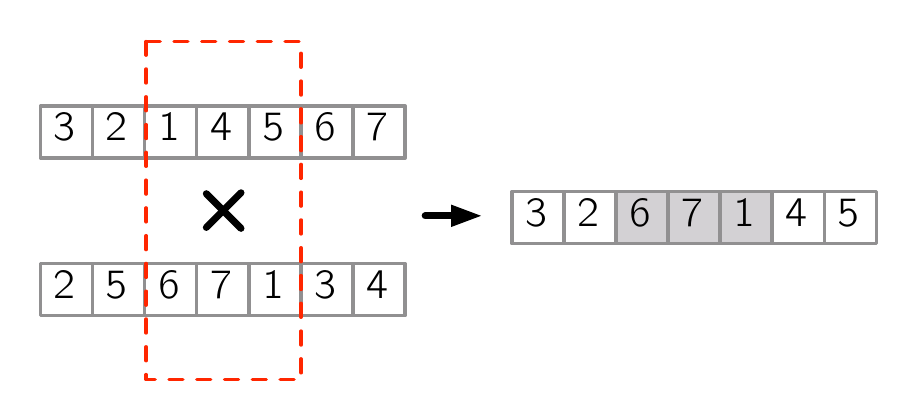}{This diagram shows the $OX$ crossover operator being applied to two tours from a \TSP. A child is produced by taking customers at position $3, 4, 5$ from the second parent and injecting these into the same position in the first parent, removing any duplicates. This leaves customers $4,5$ unrouted, so they are reinserted back into the child in the order that they appear in the first parent.}{fig:ox}{0.75}

Another common crossover operator is $EAX$. $EAX$ was originally designed for the \TSP\ but has been adapted to the \VRP\ by~\cite{Nagata:2007}. $EAX$ operates using the following process:

\begin{enumerate}
    \item Combine the two candidate solutions into a single graph by merging each solution's edge sets.
    \item Create a partition set of the graph's cycles by alternately selecting an edge from each graph.
    \item Randomly select a subset of the cycles.
    \item Generate a (incomplete) child by taking one of the parents and removing all edges from the selected subset of cycles, then add back in the edges from the parent that wasn't chosen. 
    \item Not all cycles in the child are connected to the route. Repair them by iteratively merging the disconnected cycles to the connected cycles.
\end{enumerate}

\picscl{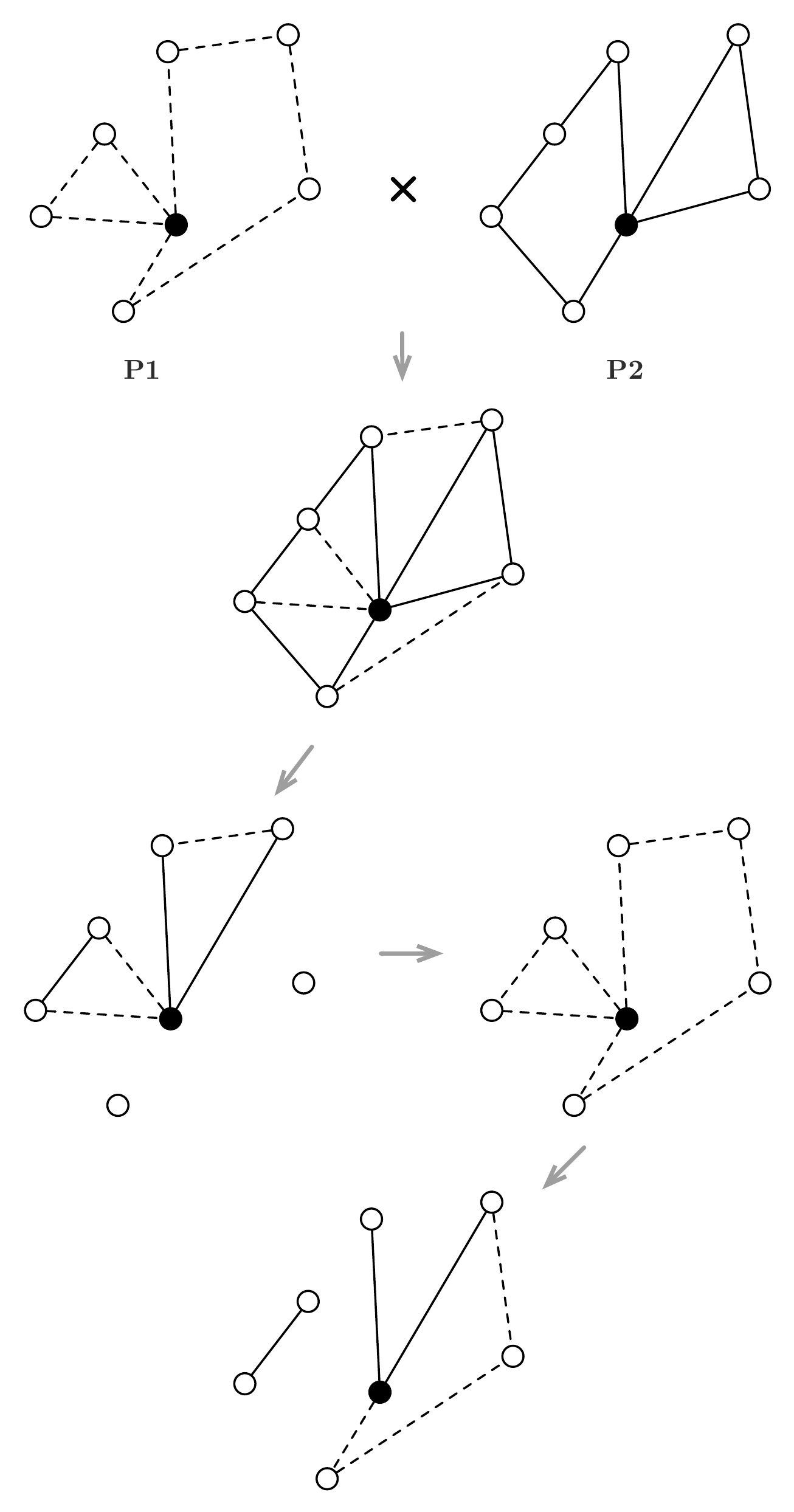}{This diagram shows an example of of $EAX$ being applied on two parent solutions, $P1$ and $P2$. The parents are first merged together. Then a new graph is created by selecting alternate edges from each parent $P1$, $P2$. A subset of cycles are then taken and applied to $P1$, such that any edges from $P1$ are removed. The child solution produced is infeasible (it contains broken routes). These would need to be repaired.}{fig:eax}{0.40}

An alternative and interesting approach found in the literature is to instead encode a set of operations and parameters that are fed to another heuristic, that in turn produces a candidate solution. A well known example of this approach was suggested in~\cite{BW:1993} which encoded an ordering of the customers. The ordering is then fed into an insertion heuristic to produce the actual candidate solutions.

An influential result that uses Genetic Algorithms to solve \VRPTW\ is given in~\cite{TNJ:1991} with their GIDEON algorithm. GIDEON uses an approach inspired by the Sweep method (an overview of the Sweep method is provided in Section~\ref{subsec:tph}). It builds routes by sweeping a ray, centered at the depot, clockwise around the geographic space enclosing the customer's locations. Customers are collected into candidate routes based on a set of parameters that are refined by the Genetic Algorithm. GIDEON uses the Genetic Algorithm to evolve the parameters used by the algorithm, rather than to operate on the problem directly. Finally, GIDEON uses a local search method to optimise customers within each route, making use of the $\lambda$-interchange operator (A description of this operator is provided in Section~\ref{subsec:simulatedannealing}).

Generally speaking, Genetic Algorithms are not as competitive as other meta-heuristics at solving the \VRP. However, more recently there have been two very promising applications of Genetic Algorithms being used to solve the \VRP. Nagata~\cite{Nagata:2007} has adapted the $EAX$ operator for use with the \VRP. And Berger and Barkaoui have presented a Hybrid Genetic Algorithm called HGA-VRP in~\cite{BM:2003}. HGA-VRP adapts a construction heuristic for use as a crossover operator. The basic premise is to select a set of routes from each parent that are located close to one another. Customers are then removed from one parent and inserted into the second using an operation inspired by Solomon's construction heuristic for \VRPTW~\cite{Solomon:1987}.

Both methods have reached the best known solution for a number of the classic \VRP\ benchmark instances by Christofides, Mingozzi and Toth~\cite{CMT:1981} and are competitive with the best Tabu Search methods.

\subsection{Tabu Search}
\label{sec:ts}

Tabu Search follows the general approach shared by many meta-heuristics; it iteratively improves a candidate solution by searching for improvements within the current solution's neighbourhood. Tabu search starts with a candidate solution, which may be generated randomly or by using another heuristic. Unlike Simulated Annealing, the best improvement within the current neighbourhood is always taken as the next move. This introduces the problem of cycling between candidate solutions. To overcome this Tabu Search introduces a list of solutions that have already been investigated and are forbidden as next moves (hence its name).

The first instance of Tabu Search being used for \VRP\ is by Willard~\cite{Willard:1989}. Willard's approach made use of the fact that \VRP\ instances can be transformed into \MTSP\ instances and solved. The algorithm uses a combination of simple vertex exchange and relocation operations. Although opening the door for further research, its results weren't competitive with the best classic heuristics. 

Osman gives a more competitive use of Tabu Search in~\cite{Osman:1993}. As with his Simulated Annealing method he makes use of the $\lambda$-interchange operation to define the search neighbourhood. Osman provides two alternative methods to control how much of the neighbourhood is searched for selecting the next move: Best-Improvement (BI) and First-Improvement (FI). Best-Improvement searches the entire neighbourhood and selects the move that is the most optimal. First-Improvement searches only until a move is found that is more optimal than the current position. Osman's heuristic produced competitive results that outperformed many other heuristics. However, it has since been refined and improved upon by newer Tabu Search methods.

Toth and Vigo introduced the concept of Granular Tabu Search (GTS)~\cite{GHL:1998}. Their method makes use of a process that removes moves from the neighbourhood that are unlikely to produce good results. They reintroduce these moves back into the process if the algorithm is stuck in a local minimum. Their idea follows from an existing idea known as Candidate Lists. Toth and Vigo's method has produced many competitive results.

Taillard has provided one of the most successful methods for solving the \VRP\ in his Tabu Search method in~\cite{Taillard:1993}. Talliard's Tabu Search uses Or's $\lambda$-interchange as its neighbourhood structure. It borrows two novel concepts from~\cite{GHL:1994}: the use of a more sophisticated tabu mechanism, where the duration (or number of iterations) that an item is tabu for is chosen randomly; and a diversification strategy, where vertices that are frequently moved without giving an improvement are penalised. A novel aspect of Taillard's algorithm is its decomposition of the problem into sub-problems. Each problem is split into regions using a simple segmentation of the region centred about the depot (Taillard also provides an alternative approach for those problems where the customers are not evenly distributed around the depot). From here each subproblem is solved individually, with customers being exchanged between neighbouring segments periodically. Taillard observes that exchanging customers beyond geographically neighbouring segments is unlikely to produce an improvement, so these moves are safely ignored. Taillard's method has produced some of the currently best known results for the standard Christofides, Mingozzi and Toth problem sets~\cite{CMT:1981}. 

\subsection{Large Neighbourhood Search}

Large Neighbourhood Search (commonly abbreviated to LNS) was recently proposed as a heuristic by Shaw~\cite{Shaw:1998}. Large Neighbourhood Search is a type of heuristic belonging to the family of heuristics known as Very Large Scale Neighbourhood search (VLSN)\footnote{LNS is somewhat confusingly named given that it a type of VLSN, and not a competing approach.}. Very Large Scale Neighbourhood search is based on a simple premise; rather than searching within a neighbourhood of solutions that can be obtained from a single (and typically quite granular) operation, such as 2-opt, it might be profitable to consider a much broader neighbourhood---a neighbourhood of candidate solutions that are obtained from applying many simultaneous changes to a candidate solution. What distinguishes these heuristics from others is that the neighbourhoods under consideration are typically exponentially large, often rendering them infeasible to search. Therefore much attention is given to providing methods that can successfully traverse these neighbourhoods. 

Large Neighbourhood Search uses a Destroy and Repair metaphor for how it searches within its neighbourhood. Its basic operation is as follows:

\begin{algorithm}[H]
   \caption{Large Neighbourhood Search}
   $x$ = an initial solution\\
   \While{termination condition not met}{
      $x^t$ = $x$\\
      $destroy(x^t)$ \\
      $repair(x^t)$ \\
      \If {$x_t$ better than current solution}{
         $x = x_t$
      }
   }
   \KwResult{ $x$ }
\end{algorithm}

It starts by selecting a starting position. This can be done randomly or by using another heuristic. Then for each iteration of the algorithm a new position is generated by destroying part of the candidate solution and then by repairing it. If the new solution is better than the current solution, then this is selected as the new position. This continues until the termination conditions are met. Large Neighbourhood Search can be seen as being a type of Very Large Scale Neighbourhood search because at each iteration the number of neighbouring solutions is exponentially large, based on the number of items removed (i.e.~destroyed).

Obviously the key components of this approach are the functions used to destroy and repair the solution. Care must be given to how these functions are constructed. They must pinpoint an improving solution from a very large neighbourhood of candidates, while also providing enough degrees of freedom to escape a local optimum.

Empirical evidence in the literature shows that even surprisingly simple destroy and repair functions can be effective~\cite{Shaw:1998}~\cite{Ropke:2005}. In applications of Large Neighbourhood Search for \VRP\ a pair of simple operations are commonly used (often alongside more complex ones too) for the destroy and repair functions. Specifically, the solution is destroyed by randomly selecting and removing $n$ customers. It is then repaired by finding the least cost reinsertion points back into the solution of the $n$ customers.

Shaw applied Large Neighbourhood Search to \VRP\ in his original paper introducing the method~\cite{Shaw:1998}. In this he introduced a novel approach for his destroy and repair functions. The destroy function removes a set of `related' customers. He defines a related customer to be any two customers that share a similar geographic location, that are sequentially routed, or that share a number of similar constraints (such as overlapping time windows if time constraints are used). The idea of removing related customers, over simply removing random customers, is that related customers are more likely to be profitably exchanged---or stated another way, unrelated customers are more likely to be reinserted back in their original positions. Shaw's repair function makes use of a simple branch and bound method that finds the minimum cost reinsertion points within the partial solution. His results were immediately impressive and reached many of the best known solutions on the Christofides, Mingozzi and Toth problems~\cite{CMT:1981}.

More recently Ropke proposed an extension to the basic Large Neighbourhood Search process in~\cite{Ropke:2005}. His method adds the concept of using a collection of destroy and repair functions, rather than using a single pair. Which function to use is selected at each iteration based on its previous performance. In this way the algorithm adapts itself to use the most effective function to search the neighbourhood. 

Ropke makes use of several destroy functions. He uses a simple random removal heuristic, Shaw's removal heuristic, and a \emph{worst} removal heuristic, which removes the most costly customers (in terms of that customer's contribution to the route's overall cost). Likewise, he makes use of several different insertion functions. These include a simple greedy insertion heuristic, and a novel insertion method he calls the `regret heuristic'. Informally, the regret heuristic reinserts those customers first who are most impacted (in terms of increased cost) by not being inserted into their optimum positions. Specifically, let $U$ be the set of customers to be reinserted and let $x_{ik}$ be a variable that gives the $k$'th lowest cost for inserting customer $i \in U$ into the partial solution. Now let $c_i^* = x_{i2} - x_{i1}$, in other words the cost difference between inserting customer $i$ into its second best position and its first. Now in each iteration of the repair function choose a customer that maximises:
\[
   \operatorname*{max}_{i \in U} c_i^*
\]

Ropke presents a series of results that show that his Large Neighbourhood Search is very competitive for solving the \VRP\ and its related problems (i.e.~\VRPTW, \PDPTW, and \DARP). Considering that Large Neighbourhood Search was only proposed in 1998, it has been very successful. In a short space of time it has attracted a large amount of research and has produced some of the most competitive results for solving the \VRP.

\section{Swarm Intelligence}
\label{sec:si}

A recent area of research is in producing heuristics that mimic certain aspects of swarm behaviour. Probably the most well known heuristics in this family are Particle Swarm Optimisation (PSO) and Ant Colony Optimisation (ACO). Real life swarm intelligence is interesting to combinatorial optimisation researchers as it demonstrates a form of emergent intelligence, where individual members with limited reasoning capability and simple behaviours, are still able to arrive at optimal solutions to complex resource allocation problems.

In the context of combinatorial optimisation, these behaviours can be mimicked and exploited. Algorithms that make use of this approach produce their solutions by simulating behaviour across a number of agents, who in themselves, typically only perform rudimentary operations. A feature of this class of algorithms is the ease with which they can be parallelised, making them more easily adaptable to large scale problems.

Swarm Intelligence algorithms have been employed to solve a number of problems. We look at two examples here, Ant Colony Optimisation and the Bees Algorithm, which this thesis makes use of.

\subsection{Ant Colony Optimisation}

Ant Colony Optimisation is inspired by how ants forage for food and communicate promising sites back to their colony. Real life ants initially forage for food randomly. Once they find a food source they return to the colony and in the process lay down a pheromone trail. Other ants that then stumble upon the pheromone trail follow it with a probability dependent on how strong (and therefore how old) the pheromone trail is. If they do follow it and find food, they then return to the colony, thus also strengthening the pheromone trail. The strength of the pheromone trail reduces over time meaning that younger and shorter pheromone trails, that do not take as long to traverse, attract more ants.  

\picscl{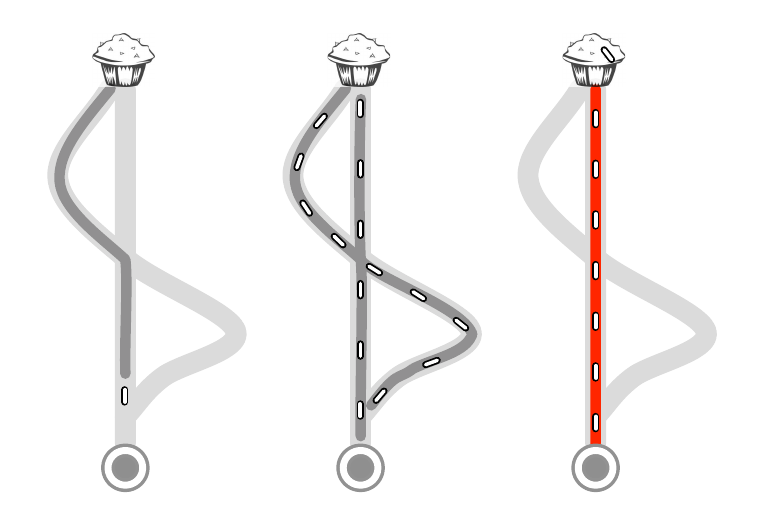}{This diagram depicts how ants make use of pheromone trails to optimise their exploitation of local food sources.}{fig:aco}{1}

Ant Colony Optimisation mimics this behaviour on a graph by simulating ants marching along a graph that represents the problem being solved. The basic operation of the algorithm is as follows: 

\begin{algorithm}[H]
   \caption{Ant Colony Optimisation}
   \KwData{A graph representing the problem}
   \While{termination condition not met}{
      positionAnts()\\
      \While{solution being built}{
         marchAnts()\\
      }
      updatePheromones()\\
   }
\end{algorithm}
 
At each iteration of the algorithm the ants are positioned randomly within the graph. The ants are then stochastically marched through the graph until they have completed a candidate solution (in the case of a \TSP\ this would be a tour of all vertices). At each stage of the march each ant selects their next edge based on the following probability formula:

\[
   p_{ij}^k = \frac{ [\tau_{ij}^{\alpha}] [\eta_{ij}^{\beta}] }{ \sum_{l \in N^k} [\tau_{il}^{\alpha}] [\eta_{il}^{\beta}] }
\]

Where $p_{ij}^k$ is the probability that ant $k$ will traverse edge \edge{i,j}, $N^k$ is the set of all edges that haven't been traversed by ant $k$ yet, $\tau$ is the amount of pheromone that has been deposited at an edge, $\eta$ is the desirability of an edge (based on a priori knowledge specific to the problem), and $\alpha$ and $\beta$ are global parameters that control how much influence each term has.

Once the march is complete and a set of candidate solutions have been constructed (by each ant, $k$), pheromone is deposited on each edge using the following equation:

\[
   \tau_{ij} = (1 - \rho) \tau_{ij} + \sum_{k=1}^m \Delta \tau_{ij}^k
\]

Where $0 < \rho \le 1$ is the pheromone persistence, and $\Delta \tau_{ij}^k$ is a function that gives the amount of pheromone deposited by ant $k$. The function is defined as:

\[
   \Delta \tau_{ij}^k = \left\{
   \begin{array}{l l}
     1 / C^k & \quad \text{if edge $\edge{i,j}$ is visited by ant $k$} \\
     0       & \quad \text{otherwise} \\
   \end{array} \right.
\]

Where $C^k$ represents the total distance travelled through the graph by ant $k$. This ensures that shorter paths result in more pheromone being deposited.

As an example of Ant Colony Optimisation's use in combinatorial problems, we show how it can be applied to the \TSP. We build a weighted graph with $i \in V$ representing each city to be visited and $\edge{i,j} \in E$ and $w_{ij}$ representing the cost of travel between each city. Then at each step of the iteration we ensure that the following constraints are met:

\begin{itemize}
   \item Each city is visited at most once.
   \item We set $\eta_{ij}$ to be equal to $w_{ij}$.
\end{itemize}

When the Ant Colony Optimiser starts, it positions each ant at a randomly selected vertex (i.e.~city) within the graph. Each step of an ant's march then builds a tour through the cities. Once an ant has completed a tour it serves as a candidate solution for the \TSP. Initially the solutions will be of low quality, so we use the length of the tours to ensure that more pheromone is deposited on the shorter tours. At the end of $n$ iterations the ants will have converged on a near optimal solution (but like all meta-heuristics there's no guarantee that this will be the global optimum).

\picscl{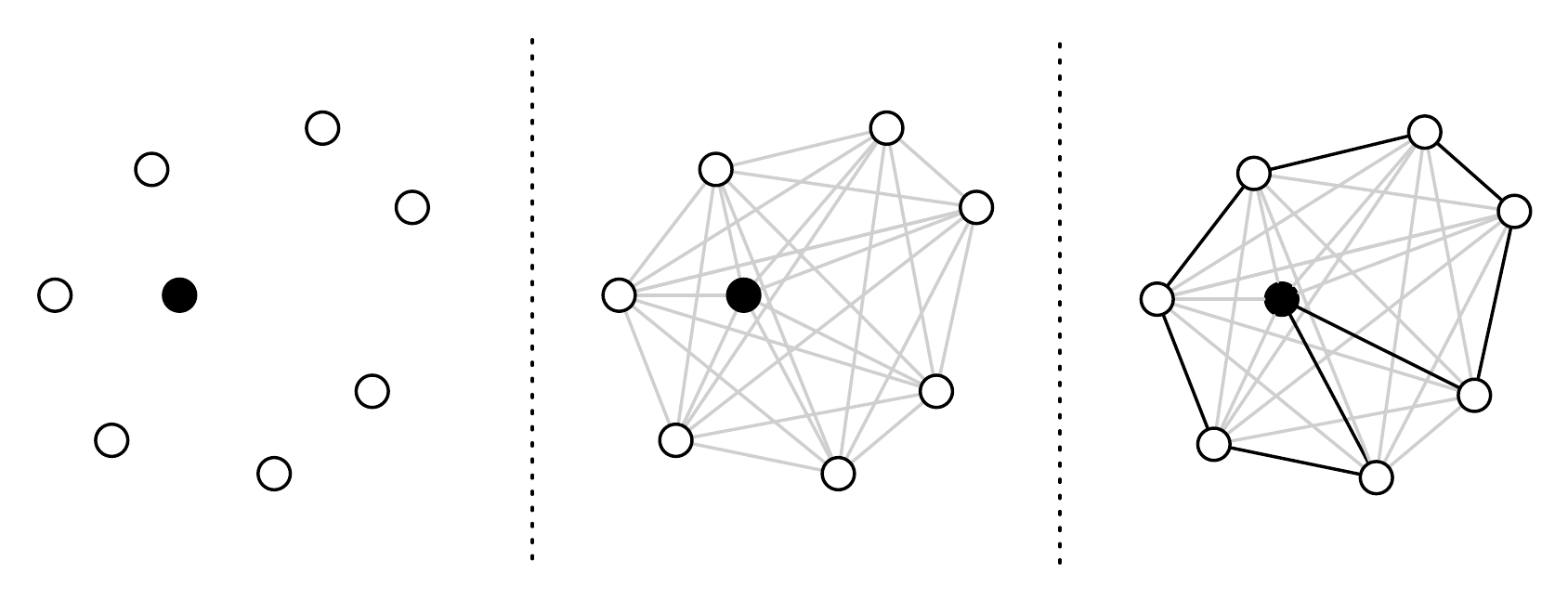}{Shown is an example of ACO being used to solve a \TSP. Initially the ants explore the entire graph. At the end of each iteration the more optimal tours will have more pheromone deposited on them, meaning in the next iteration the ants are more likely to pick these edges when constructing their tour. Eventually the ants converge on a solution.}{fig:acotsp}{0.66}
 
Ant Colony Optimisation has been applied to \VRP\ by Bullnheimer, Hartl, and Strauss in~\cite{BHS:1999A}~\cite{BHS:1999B}. They adapted the straightforward implementation used for the \TSP, detailed in the preceding discussion, by forcing the ant to create a new route each time it exceeds the capacity or maximum distance constraint. They also use a modified edge selection rule that takes into account the vehicle's capacity and its proximity to the depot. Their updated rule is given by:

\[
   p_{ij}^k = \frac{ [\tau_{ij}^{\alpha}] [\eta_{ij}^{\beta}] [s_{ij}] [\kappa_{ij}]  }{ \sum_{l \in N^k} [\tau_{il}^{\alpha}] [\eta_{il}^{\beta}] [s_{il}] [\kappa_{il}] }
\]

Where $s$ represents the proximity of customers $i, j$ to the depot, and $\kappa = (Q^i + q^j) / Q$---$Q$, giving the maximum capacity, $Q^i$ giving the capacity already used on the vehicle, and $q^j$ is the additional load to be added. $\kappa$ influences the ants to take advantage of the available capacity. 

Bullnheimer et al.'s implementation of Ant Colony Optimisation for \VRP\ produces good quality solutions for the Christofides, Mingozzi and Toth problems~\cite{CMT:1981}, but is not competitive with the best modern meta-heuristics.

More recently Reimann, Stummer, and Doerner have presented a more competitive implementation of Ant Colony Optimisation for \VRP~\cite{RSD:2002}. Their implementation operates on a graph where $\edge{i,j} \in E$ represent the savings of combining two routes, as given by the classic Clark and Wright Savings heuristic (see Section~\ref{subsec:conheu} for more on this heuristic). Each ant selects an ordering of how the merges are applied. This implementation is reported to be competitive with the best meta-heuristics~\cite{Potvin:2009}.   

\subsection{Bees Algorithm}
\label{subsec:beesalgorithm}

Over the last decade, and inspired by the success of Ant Colony Optimisation, there have been a number of algorithms proposed that aim exploit the collective behaviour of bees. This includes: \emph{Bee~Colony~Optimisation}, which has been applied to many combinatorial problems, \emph{Marriage~in~Honey~Bees~Optimization} (MBO) that has been used to solve propositional satisfiability problems, \emph{BeeHive} that has been used for timetabling problems, the \emph{Virtual~Bee~Algorithm} (VBA) that has been used for function optimisation problems, \emph{Honey-bee~Mating~Optimisation} (HBMO) that has been used for cluster analysis, and finally, the \emph{Bees~Algorithm} that is the focus of this thesis. See~\cite{LJDS:2009} for a bibliography and high level overview on many of these algorithms. 

The Bees Algorithm was first proposed in~\cite{PGKORZ:2005}. It is inspired by the foraging behaviour of honey bees. Bee colonies must search a large geographic area around their hive in order to find sites with enough pollen to sustain a hive. It is essential that the colony makes the right choices in which sites are exploited and how much resource is expended on a particular site. They achieve this by sending scout bees out in all directions from the hive. Once a scout bee has found a promising site it then returns to the hive and recruits hive mates to forage at the site too. The bee does this by performing a \emph{waggle} dance. The dance communicates the location and quality of the site (i.e.~fitness). Over time, as more bees successfully forage at the site, more are recruited to exploit the site---in this aspect bee behaviour shares some similarities with ant foraging behaviour. 

\picscl{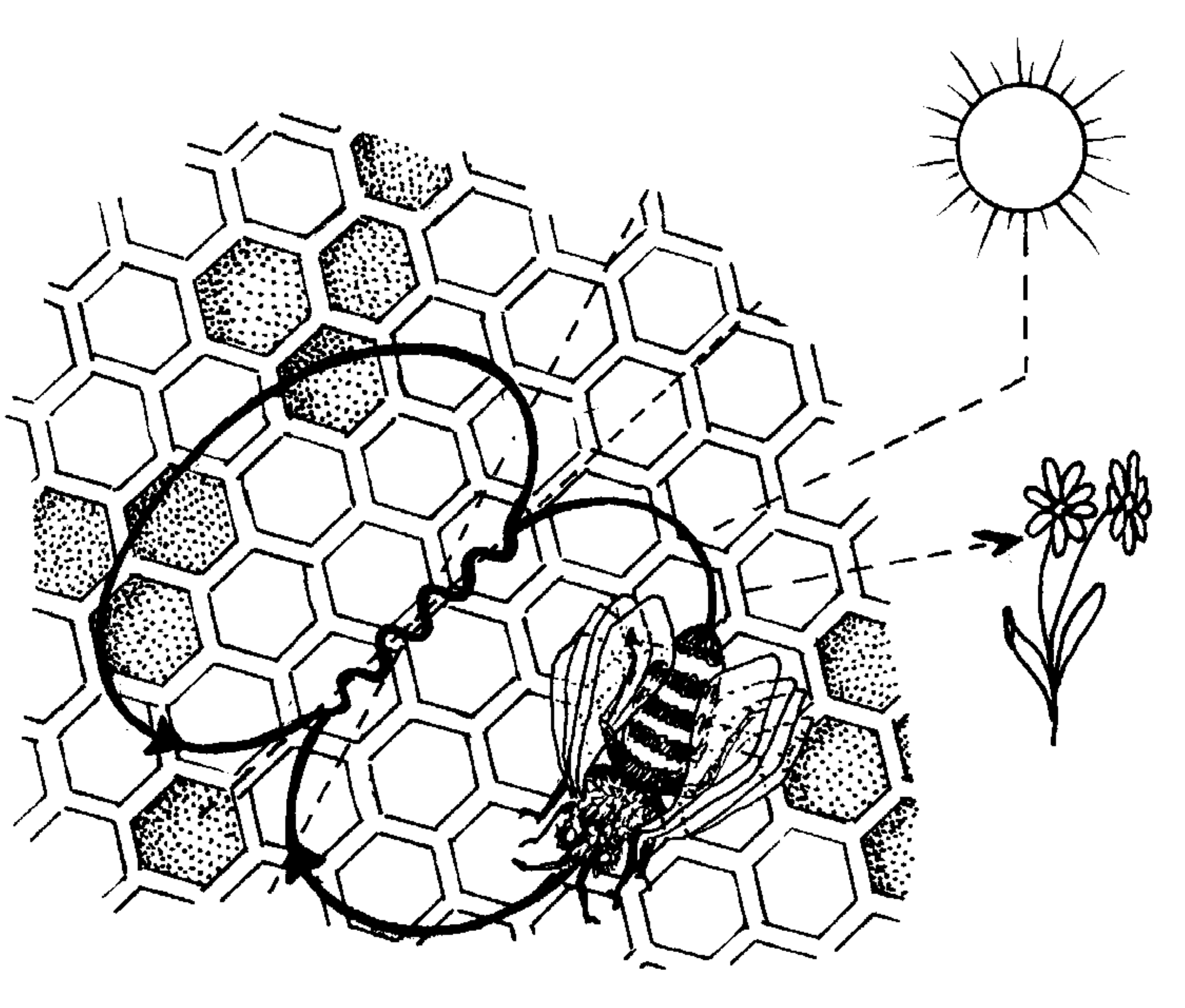}{Shown is the \emph{waggle} dance performed by a honey bee (image courtesy of~\cite{biodidac:website}). The direction moved by the bee indicates the angle that the other bees must fly relative to the sun to find the food source. And the duration of the dance indicates its distance.}{fig:waggle}{0.12}

Informally the algorithm can be described as follows. Bees are initially sent out to random locations. The fitness of each site is then calculated. A proportion of the bees are reassigned to those sites that had the highest fitness values. Here each bee searches the local neighbourhood of the site looking to improve the site's fitness. The remainder of the bees are sent out scouting for new sites, or in other words, they are set to a new random position. This process repeats until one of the sites reaches a satisfactory level of fitness---or a predetermined termination condition is met.

More formally, the algorithm operates as follows:

\begin{algorithm}[H]
   \caption{Bees Algorithm}
   $B = \set{b_1, b_2,\ldots, b_n}$\\
   setToRandomPosition($B$)\\
   \While{termination condition not met}{
      sortByFitness($B$) \\
      $E = \set{b_1, b_2,\ldots,b_e}$ \\
      $R = \set{b_{e+1}, b_{e+2},\ldots,b_{m}}$ \\
      searchNeighbourhood($E \union \set{c_1,\ldots,c_{nep}}$) \\
      searchNeighbourhood($R \union \set{d_1,\ldots,d_{nsp}}$) \\
      setToRandomPosition($B - (E \union R)$) \\
   }
\end{algorithm}

$B$ is the set of bees that are used to explore the search space. Initially the bees are set to random positions. Function $sortByFitness$ sorts the bees in order of maximum fitness. It then proceeds by taking the $m$ most promising sites found by the bees. It does this by partitioning these into two sets, $E, N \subset B$. $E$ is the first $e$ best sites, and represents the so called \emph{elite} bees. $N$ is the $m - e$ next most promising sites. The $searchNeighbourhood$ function explores the neighbourhood around a provided set of bees. Each site in $E$ and $N$ is explored. $nep$ bees are recruited for the search of each $b \in E$, and $nsp$ are recruited for the search of each $b \in N$. In practice this means that $nep$ and $nsp$ number of positions are explored within the neighbourhoods of $E$ and $N$'s sites, respectively. These moves are typically made stochastically, but it is possible for a deterministic approach to be used too. The remaining $n - m$ bees (in other words, those not in $E$ and $N$) are set to random positions. This is repeated until the termination condition is met, which may be a running time threshold or a predetermined fitness level. 

The advantage promised by the Bees Algorithm over other meta-heuristics is its ability to escape local optima and its ability to navigate search topologies with rough terrain (such as in Figure~\ref{fig:rtst}). It achieves this by scouting the search space for the most promising sites, and then by committing more resources to the exploration of those sites that produce better results.

\picscl{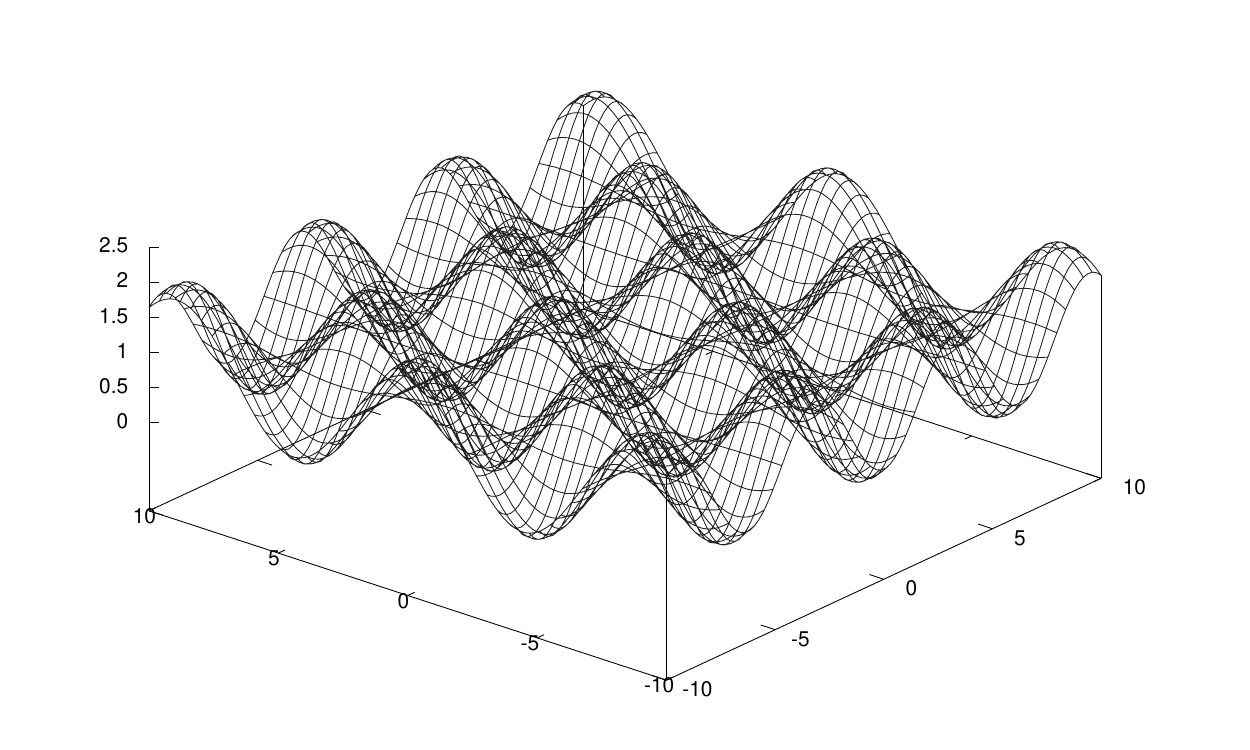}{Shown is a search space with many valleys and hills. These search spaces provide a challenge to meta-heuristic approaches as there are many local minima and maxima to get caught in. The Bees Algorithm ameliorates this by searching in many different areas simultaneously.}{fig:rtst}{0.66}

The Bees Algorithm has been applied to manufacturing cell formation, training neural networks for pattern recognition, scheduling jobs for a production machine, data clustering, and many others areas. See~\cite{beesalg} for more examples and a comprehensive bibliography. However, to the best of our knowledge, the Bees Algorithm hasn't been adapted for the Vehicle Routing Problem until now.


\chapter{Problem Definition}
\label{chap:pd}

In this chapter we provide a formal definition of the \VRP\ and briefly describe the variant problems that have arisen in the literature. The Capacitated Vehicle Routing Problem (\CVRP) is the more correct name for the \VRP\ that distinguishes it from its variants. We start in Section~\ref{sec:capacitatedvehicleroutingproblem} by providing a formal definition of the \CVRP. We formulate it as an integer linear programming problem, as has become standard in the \VRP\ literature. And follow this in Section~\ref{sec:variants} by an overview of the \VRP\ variants that are commonly used.

\section{Capacitated Vehicle Routing Problem}
\label{sec:capacitatedvehicleroutingproblem}

We formulate the \CVRP\ here as an integer linear programming problem. Although it is possible to solve the \CVRP\ using an integer programming solver, this is uncommon in practice as the best solvers are still only able to solve for small problem sizes. We provide this formulation as it has become the lingua franca of combinatorial problems.

We start the formation by specifying the variables used within it. We represent the \CVRP\ on a weighted graph, $G = (V, E)$. The vertices of the graph $V$ represent all locations that can be visited, this includes each customer location and the location of the depot. For convenience we let $v^d$ denote the vertex that represents the depot, and we denote the set of customers as $C = \set{1,2,\ldots,n}$. Thus the set of vertices is given by $V = v^d \union C$. We now let the set of edges, $E$, correspond to the valid connections between customers and connections to the depot. For the \CVRP\ all connections are possible, in other words, we set $G$ to be a clique. Each edge $\edge{i,j} \in E$ has a corresponding cost $c_{ij}$. We let the cost be the euclidian distance between the two locations $c_{ij} = \sqrt{(x_j - x_i)^2 + (y_j -  y_i)^2}$. Where $x_i$ and $y_i$ for $i \in V$ represent the coordinates of the customer's location. 

We use $K$ to denote the set of vehicles that are used to visit customers, such that $|K| = m$ and $m$ is the maximum number of vehicles allowed. We define $q$ and $t$ to be the maximum capacity and the maximum work duration, respectively, allowable for a vehicle. The demand (i.e.~required capacity) for each customer is denoted by $d_i, i \in C$. Likewise, we denote the service time required by each customer as $t_i, i \in C$. We then use the decision variable $X_{ij}^k$ to denote if a particular edge $\edge{i,j} \in E$ is traversed by vehicle $k \in K$, in other words, $k$ travels between customers $i,j \in C$. Where this is true we let $X_{ij}^k = 1$, and $X_{ij}^k = 0$ where it is not. We use $u_i, i \in C$ as a sequencing variable that gives the position of customer $i$ within the route of the vehicle that visits it.

We are now able to define the problem as follows:
\begin{align}\notag
\intertext{Minimise:}
   &\sum_{k \in K} \sum_{\edge{ij} \in E} c_{ij}X_{ij}^k \tag{3.1}\label{PF:1}
\intertext{Subject to:} 
   &\sum_{k \in K} \sum_{j \in V} X_{ij}^k = 1
      &&\quad \forall i \in C \tag{3.2}\label{PF:2}\\
   &\sum_{i \in C} d_i \sum_{j \in C} X_{ij}^k \leq q
      &&\quad \forall k \in K \tag{3.3}\label{PF:3}\\
   &\sum_{i \in C} t_i \sum_{j \in C} X_{ij}^k + \sum_{\edge{ij} \in E} c_{ij}X_{ij}^k \leq t
      &&\quad \forall k \in K \tag{3.4}\label{PF:4}\\
   &\sum_{j \in V} X_{v^dj}^k = 1
      &&\quad \forall k \in K \tag{3.5}\label{PF:5}\\
   &\sum_{j \in V} X_{jv^d}^k = 1
      &&\quad \forall k \in K \tag{3.6}\label{PF:6}\\
   &\sum_{i \in V} X_{ic}^k - \sum_{j \in V} X_{cj}^k = 0
      &&\quad \forall c \in C \mbox{ and } \forall k \in K \tag{3.7}\label{PF:7} \\
   &u_i - u_j + |V| X_{ij}^k \leq |V| - 1
      &&\quad \forall \edge{i,j} \in E - v^d \mbox{ and } \forall k \in K \tag{3.8}\label{PF:8} \\
   &X_{ij}^k \in \set{0, 1}
      &&\quad \forall \edge{i,j} \in E \mbox{ and } \forall k \in K \tag{3.9}\label{PF:9}
\end{align}

The objective function~\eqref{PF:1} minimises the costs $c_{ij}$. Constraint~\eqref{PF:2} ensures that each customer can only be serviced by a single vehicle. Constraint~\eqref{PF:3} enforces the capacity constraint; each vehicle cannot exceed its maximum vehicle capacity $q$. Likewise Constraint~\eqref{PF:4} enforces the vehicle's work duration constraint. A vehicle's work duration is the sum of its service times ($t_i, i \in C$ where customer $i$ is visited by the vehicle) and its travel time. By convention the travel time is taken to be equal to the distance traversed by the vehicle, which in turn is equal to the costs, $c_{ij}$, of the edges it traverses. Constraints~\eqref{PF:5} and~\eqref{PF:6} ensure that each vehicle starts at the depot and finishes at the depot, and that they do this exactly once. Constraint~\eqref{PF:7} and Constraint~\eqref{PF:8} are flow constraints that ensure that the number of vehicles entering a customer is equal to the number of vehicles leaving, and that sub-tours are eliminated. Lastly, Constraint~\eqref{PF:9} ensures the integrality conditions.

Constraint~\eqref{PF:4}, which enforces a maximum vehicle work duration, $t$, is often left out of the traditional \CVRP\ formation but is included here as it is present in the problem instances we use for benchmarks in Chapter~\ref{chap:results}.

\section{Variants}
\label{sec:variants}

In this section we provide an overview of the common variations of the \VRP\ that are used. These variations have arisen from real world vehicle routing scenarios, where the constraints are often more involved than is modelled in the \CVRP. 

\subsection{Multiple Depot Vehicle Routing Problem}

A simple extension to the \CVRP\ is to allow each vehicle to start from a different depot. Part of the problem now becomes assigning customers to depots, which in itself is a hard combinatorial problem. The \CVRP\ formation can easily be relaxed to allow this. There are two variations of the problem. One constrains each vehicle to finish at the same depot that it starts from. The other allows vehicles to start and finish at any depot, as long as the same number of vehicles return to the depot as left from it.

\subsection{Vehicle Routing with Time Windows}

The Vehicle Routing Problem with Time Windows (\VRPTW) adds the additional constraint to the classic \VRP\ that each customer must be visited within a time window specified by the customer. More formally, for \VRPTW\ each customer $i \in C$ also has a corresponding time window $\seq{a_i, b_i}$ in which the goods must be delivered. The vehicle is permitted to arrive before the start time, $a_i$. However, in this case the vehicle must wait until time $a_i$ adding to the time it takes to complete the route. However, it is not permitted for the job to start after time $b_i$.

An additional constraint is added to the formation of \CVRP\ to ensure that time window constraints are met: $a_i \leq S_i^k \leq b_i$ where the decision variable $S_i^k$ provides the time that each vehicle $k \in K$ arrives at customer $i \in V$.

\subsection{Pickup and Delivery Problem}

The Pickup and Delivery Problem (\PDP) generalises the \VRP. In this problem goods are both picked up and delivered by the vehicle along its route. The vehicle's work now comes in two flavours: pickup jobs, $P = \set{p_1,p_2,\ldots,p_k}$, and delivery jobs, $D = \set{d_1,d_2,\ldots,d_l}$, such that $C = P \union D$. Additional constraints are added to the \CVRP\ formation to ensure that:

\begin{enumerate}
   \item Pickup and deliver jobs are completed by the same vehicle, that is $p_i \in R_k \implies d_i \in R_k$ where $R_k$ represents a sequence of jobs undertaken by a vehicle $k$.

   \item The pickup job, $p_i$, appears before its corresponding delivery job, $d_i$, in the sequence of jobs undertaken by a vehicle.

   \item The vehicles capacity is not exceeded as goods are loaded and unloaded from it. This requires the use of an intermediate variable, $y^k_i, i \in V, k \in K$ that represents the load of vehicle $k$ at customer $i$. It adds constraints: $y^k_0 = 0$, $X_{ij}^k = 1 \implies y^k_j = y^k_i + d_i$, and $\sum_{i \in V} y^k_i \leq q$ for all $k \in K$, to enforce this.
   
\end{enumerate}

There is also a variation on \PDP\ that adds time windows, called \PDPTW. In this case the extra constraints from the \VRPTW\ problem are merged with those given here. \PDP\ is a much harder problem computationally than \CVRP, as its extra constraints add new dimensions to the problem. Because of its complexity \PDP\ has only been actively researched in the last decade.


\chapter{Algorithm}
\label{chap:algorithm}

This chapter provides a detailed description of the Enhanced Bees Algorithm, the algorithm developed for this thesis, and its operation. We start by reviewing the objectives that the algorithm was designed to meet in Section~\ref{sec:objectives}. In Section~\ref{sec:problemrepresentation} we provide a description of how the algorithm internally represents the \VRP\ problem and its candidate solutions. Next in Section~\ref{sec:enhancedbeesalgorithm} we provide a detailed description of the operation of the algorithm. Finally, in Section~\ref{sec:searchneighbourhood} we describe the neighbourhood structures that are used by the algorithm to define its search space.

\section{Objectives}
\label{sec:objectives}

The Enhanced Bees Algorithm was built for use in a commercial setting. It was developed as part of a New Zealand Trade and Enterprise grant for the company \emph{vWorkApp~Inc.}'s scheduling and dispatch software. Accordingly, different objectives were aimed for with its design (such as runtime performance) than are typically sought in the \VRP\ literature. The algorithm's objectives, in order of priority, are as follows:

\begin{enumerate}
   \item Ensure that all constraints are met. Specifically that the route's maximum duration is observed. 
   
   \item Have a good runtime performance. It is more desirable for the algorithm to produce a reasonable quality result quickly (within 60 seconds), than for it to produce a better result but require a longer processing time. Specifically if the algorithm could reach 5\% of the optimum value within 60 seconds then this would be sufficient. 

   \item Produce good quality results. Notwithstanding objective 2, the results produced must be close to the global optimum. 
   
   \item Have a design that lends itself to parallelisation and is able to make use of the additional processing cores available within modern hardware.
   
\end{enumerate}

\section{Problem Representation}
\label{sec:problemrepresentation}

The Enhanced Bees Algorithm represents the problem in a direct and straightforward manner. It directly manipulates a candidate solution $\schd$, where $\schd$ is a set of routes $R \in \schd$, and each route contains an ordered sequence of customers $v_i \in R$ starting and ending at the depot vertex $v^d$.

\picscl{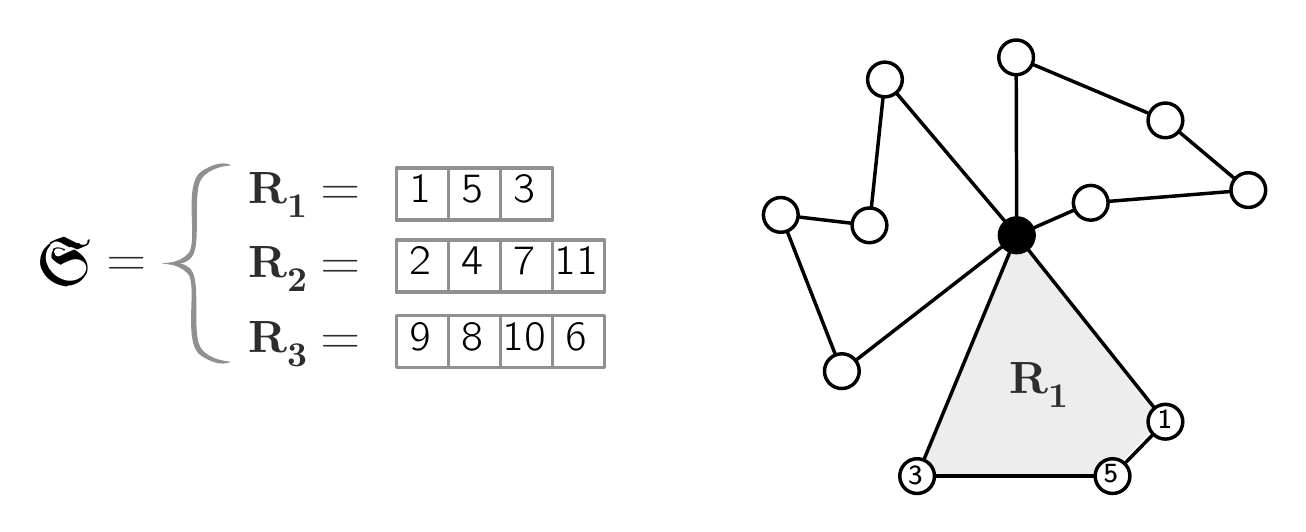}{Shown is an example of a simple \VRP\ candidate solution as represented internally by the Enhanced Bees Algorithm.}{fig:representation}{0.66}

More general representations are sometimes used for meta-heuristics, as is commonly seen with Genetic Algorithms, as they allow the algorithm to be easily adapted to other combinatorial problems. However, this often comes at a cost of added complexity and inferior results\footnote{This occurs because the operators that act on the problem representation can no longer exploit information that is specific to the problem domain and must rely on general purpose operations instead.}. This algorithm was designed specifically for solving instances of the \VRP\ so a direct representation was chosen.

The algorithm makes use of a \emph{fitness} concept, common to many meta-heuristics, to describe the cost of the solution. The fitness function $f()$ includes terms for the distance (i.e.~cost) of the solution and penalties for breaking the capacity and maximum route duration constraints. The Enhanced Bees Algorithm uses penalties to encourage feasible solutions to be produced. Rather than outright barring infeasible solutions, the fitness function allows the algorithm some wriggle room to traverse through these on its way towards a feasible solution.

Specifically $f()$ is defined as follows:
\begin{align}
   & c(R) = \sum_{i \in R} c_{i, i+1} \\
   & d(R) = max\left( \sum_{i \in R} d_i - q, 0 \right)  \\
   & t(R) = max\left( \sum_{i \in R} t_i + c(R) - t, 0 \right)  \\
   & f(\schd) = \sum_{R \in \schd} (\alpha c(R) + \beta d(R) + \gamma c(R))
\end{align}

Function $c(R)$ calculates the cost (i.e distance) of a given route, and function $d(R)$ calculates how overcapacity the given route is. We define overcapacity to be how much larger the sum of the route's demands, $d_i, i \in R$, are than the stated maximum allowable capacity $q$. Likewise function $t(R)$ calculates the overtime of a given route. A route's duration is calculated as being the sum of its customer's service times, $t_i, i \in R$, and its travel time. By convention the travel time is equal to the distance of the route. Function $t(R)$ then returns how much over the maximum allowable route duration, $t$, the duration is. Lastly, the fitness function $f()$ is the weighted sum of these three terms. Parameters $\alpha$, $\beta$, and $\gamma$ are used used to control how much influence each term has on determining the candidate solution's fitness.

For the purposes of benchmarking our algorithm (see Chapter~\ref{chap:results}) we use a travel cost that is equal to the 2D Euclidian distance\footnote{Specifically, we use: $c_{ij} = \sqrt{(x_j-x_i)^2 + (y_j-y_i)^2}$} between the two points. For real life problems we have found that using a manhattan distance\footnote{Specifically, we use: $c_{ij} = (x_j-x_i) + (y_j-y_i)$} often provides superior results. This is presumably due to the manhattan distance better modelling the road system we tested on (Auckland,~New~Zealand), which although not a strict grid, is still closer to this than the Euclidian distance models.

\section{Enhanced Bees Algorithm}
\label{sec:enhancedbeesalgorithm}

Our Algorithm is based on the \emph{Bees~Algorithm} (see Section~\ref{subsec:beesalgorithm} for an overview of the standard Bees Algorithm). The Enhanced Bees Algorithm makes some changes to adapt the Bees Algorithm to the \VRP\ domain. An interesting aspect of the Bees Algorithm is that it covers a broad search area, minimising the risk of being stuck in a local optimum. It achieves this by randomly probing (or in the Bees Algorithm parlance, `scouting') many areas of the search space through its entire run. However, this approach is not well suited to hard combinatorial problems, where a newly constructed solution, let alone a randomly generated one, is often far from optimal (for instance, the Clark Wright Savings heuristic still produces solutions that are up to 15\% from the global optimum and would require many operations to get close to optimal). We have adapted the Bees Algorithm such that many of its unique characteristics, like its relative robustness, are maintained while working well with hard combinatorial problems, such as the \VRP.

The Enhanced Bees Algorithm can be summarised, at a high level, as follows:

\begin{algorithm}[H]
   \caption{Enhanced Bees Algorithm}
   $S$ = seedSites() \\

   \While{termination condition not met}{
      \For{$s_i \in S$}{
         explore($s_i$, $d$) \\
         \If{$i < \lambda$}{
            removeWorstSite \\
         }
      }
   }
\end{algorithm}

The algorithm maintains a collection of sites $S$, and each site $s_i \in S$ maintains a collection of bees, $B_i$. Each bee is a proxy to the problem domain that we are trying to solve. In our case this is the \VRP\ problem representation covered in Section \ref{sec:problemrepresentation}.

Initially each site is seeded, such that each site, $s_i \in S$, contains a collection of bees $B_i$, and each bee has a corresponding \VRP\ candidate solution, $\schd$. Each candidate solution is initialised by seeding each route with a randomly chosen customer, which is then filled out using the insertion heuristic outlined in Section~\ref{subsec:repairheuristic}. Each site is then in turn improved upon. This is achieved by iteratively exploring the neighbourhood of each site. The process used to explore each site is where the majority of the algorithm's processing takes place and where the interesting aspects of the algorithm come into play. The exploration process is covered in detail in Sections~\ref{sec:beemovement},~\ref{subsec:searchspacecoverage}, and~\ref{sec:searchneighbourhood}.

The number of sites explored is reduced over the run of the algorithm. This borrows from the idea of a cooling schedule used in Simulated Annealing. Sites are reduced using the formula:

\begin{align}
   & S = S - s_w & \text{if $i \bmod{\lambda} \equiv 0$}
\end{align}

Where $s_w$ represents the worst site, in terms of fitness, $i$ represents the current iteration of the algorithm, and $\lambda$ represents the period of iterations with which the number of sites are reduced. Once the algorithm is complete the solution, $\schd$, with the best overall fitness is returned as the answer. In the next section we review in more detail each aspect of the algorithm.

\subsection{Bee Movement}
\label{sec:beemovement}

Bees are moved around the search space to look for improvements to the collection of candidate solutions being maintained. Each bee represents a candidate solution, $\schd$, so a valid bee move is any new candidate solution $\schd'$ that can be reached within the neighbourhood of $\schd$ (see Section~\ref{sec:searchneighbourhood} for the operations under which the neighbourhood is defined).

\picscl{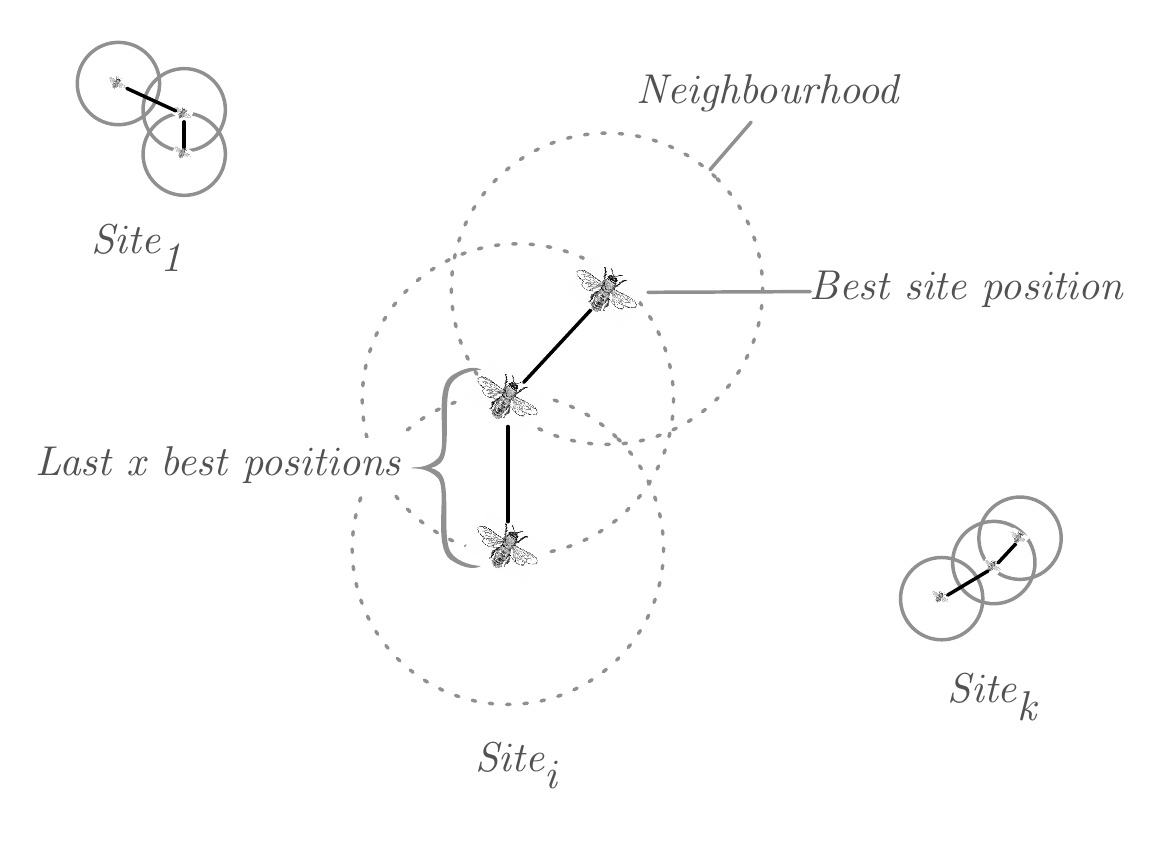}{Three sites are shown along with their neighbourhoods. Site $i$ shows in more detail that each site maintains a list of the last $\epsilon$ most promising positions for exploration.}{fig:movement}{0.85}

A feature of the Enhanced Bees Algorithm is that two Bees cannot occupy the same position. The algorithm maintains a register of the current positions occupied by each bee. We use the current fitness, $f(\schd)$, as a quick and simple representation of a bee's current position\footnote{This obviously will not work in circumstances where there is a reasonable likelihood of two candidate solutions, $\schd_i$ and $\schd_j$ having $f(\schd_i) = f(\schd_j)$. While this is not the case with the problem instances we have used in this thesis, this may need to be modified if the algorithm is to be used on more general problem instances.}. If a bee tries to occupy the same position as another bee (i.e.~they share the same candidate solution) then the bee trying to occupy that position is forced to explore the neighbourhood again and find another position. 

Enforcing the constraint that each bee must occupy a unique position has two benefits: it forces diversification between the bees and sites, hence encouraging a greater proportion of the search space to be explored; and it increases the chance of a local optimum being escaped, as a bee ensnared in the local optimum now forces the remainder of the hive to explore alternative positions. This feature has a similar intent and effect to the tabu lists used by Tabu Search.

Another feature of the Enhanced Bees Algorithm is the role that sites play in concentrating exploration on certain areas of the search space. Each site maintains a list, $M = \seq{\schd_1,\ldots, \schd_\epsilon}$, of the last $\epsilon$ best positions. Each $\schd_i \in M$ is then taken as a launching point for a site's bees to explore. $\theta$ bees are recruited for the exploration of each $\schd_i \in M$. Once all positions in $M$ have been explored then the best $\epsilon$ positions are again taken and used as the launching points for the site's next round of exploration. This exploration method has two purposes: firstly, it allows for a simple type of branching, as $\epsilon$ of the most promising positions that were traversed through on the way to the current position are also explored; secondly, it prevents cycling between promising solutions that are in close vicinity to each other.

Conversely, sites do not interact with each other, as each site maintains its own unique list of $\epsilon$ promising positions. The constraint that no two bees can occupy the same position ensures that each site covers a non overlapping area of the search space. In practice we have found that this is sufficient to encourage sites to diverge and explore distinct areas of the search space.    


\subsection{Search Space Coverage}
\label{subsec:searchspacecoverage}

As mentioned, one of the unique aspects of the Bees Algorithm is its ability to produce robust results through probing a large area of the search space. However, this doesn't work well with hard combinatorial problems, where it cannot be ascertained quickly if an area in the search space shows promise or not. 

To overcome this limitation we instead use an approach inspired by Simulated Annealing's use of a cooling schedule. Bees are initially divided equally between each site $s_i \in S$, ensuring that each site is explored equally. Then every $\lambda$ period of iterations we reduce the number of sites maintained, such that $S = S - s_w$, where $s_w$ is the site with lowest fitness. We measure each site's fitness from the fitness of its best position found to date.

This process continues until a single site remains. We show experimentally in Chapter~\ref{chap:results} that this process improves the robustness of the algorithm and produces better results overall than the standard Bees Algorithm. 

\section{Search Neighbourhood}
\label{sec:searchneighbourhood}

As already discussed, each bee seeks to improve upon its current fitness by exploring the local neighbourhood of the solution it represents. The Enhanced Bees Algorithm does this by applying a Large Neighbourhood Search (LNS) operator to its candidate solution $\schd$. The LNS operator differs from the more common \VRP\ operators in that a single operation applies many changes to the candidate solution $\schd$. This widens the neighbourhood of $\schd$ to encompass exponentially many candidate solutions. LNS navigates through the vast space it spans by selecting only those changes that have a high likelihood of improving the solution.

The LNS operation is comprised of two-phases: a destroy phase, and a repair phase. LNS's destroy phase, when used for the \VRP, typically involves removing a proportion of the customers from the solution. The Enhanced Bees Algorithm's destroy phase uses two heuristics along those lines: a somewhat intelligent heuristic that attempts to remove those customers that are more likely to be able to be recombined in a profitable way; and a simple random selection. These heuristics are covered in more detail in Section~\ref{subsec:destroyheuristic}. The second phase of LNS is used to repair the partial solution. The Enhanced Bees Algorithm uses a simple insertion heuristic that inserts the customers into those locations that have the lowest insertion cost. This heuristic is covered more formally in Section~\ref{subsec:repairheuristic}.

\subsection{Destroy Heuristic}
\label{subsec:destroyheuristic}

The Enhanced Bees Algorithm employs two destroy heuristics. The first simply selects $l$ customers randomly from a solution $\schd$ and removes these from their routes. The second is slightly more complicated and is due to Shaw~\cite{Shaw:1998}. Shaw's removal heuristic stochastically selects customers such that there is a higher likelihood of customers that are related to one another being removed. For our purposes we define \emph{related} to mean that for any two customers, $v_i, v_j \in V$, then either $v_i, v_j$ are geographically close to one another (i.e.~$c_{ij}$ is small), or they share an adjacent position within the same route, $R = \seq{\ldots, v_i, v_j,\ldots}$.

The rationale to removing related customers is that they are the most likely to be profitably exchanged with one another. Conversely, unrelated customers are more likely to be reinserted back into the same positions that they were removed from. 

\subsection{Repair Heuristic}
\label{subsec:repairheuristic}

The repair heuristic used by the Enhanced Bees Algorithm randomly selects one of the removed customers $v_j$, and calculates a cost for reinserting $v_j$ between each pair of jobs $v_i, v_k \in R_i$ for all $R_i \in \schd$ (actually not all reinsertion positions are considered, see Section~\ref{subsec:neighborhoodscope} for a description of which positions are considered). The reinsertion cost is calculated as follows:

\begin{align}
   c^*   &= c_{ij} + c_{jk} - c_{ik} \\
   cost  &= c^* + d(R') - d(R) + t(R') - t(R)
\end{align}

Where $c^*$ calculates the cost difference in terms of travel distance. $R$ and $R'$ are defined as the route before and after the customer is inserted, respectively. And functions $d(R)$ and $t(R)$ are all defined as they are in Section~\ref{sec:problemrepresentation}. The final cost is the sum of the added travel distance and the two extra penalties, if the route is now overcapacity or over its maximum duration. The algorithm selects the position with the lowest insertion cost to reinsert the customer. This is repeated until all customers are reinserted into the solution. 

The reason that customers are reinserted in a random order is that it adds a beneficial amount of noise to the heuristic. This ensures that a healthy diversity of solutions are generated from the heuristic. 

\subsection{Neighbourhood Extent}
\label{subsec:neighborhoodscope}

We use two techniques to adjust the extent of the neighbourhood being searched. The first technique that we use allows the algorithm some flexibility in selecting infeasible solutions. As can be seen from our formulation of the candidate solutions' fitness values (see Section~\ref{sec:problemrepresentation}) violations of the problem's capacity and duration constraints are penalised rather than forbidden. This allows the bees to navigate through infeasible solutions, where other aspects of that solution are sufficiently attractive enough to outweigh the penalties. However, only feasible solutions are allowed to be counted as final solutions returned by the algorithm.

The second technique that we use is to adjust the number of insertion positions considered as part of the repair heuristic. The number of insertion positions considered starts with both sides of the three closest customers and increases as the site ages. More formally, let $v_i \in V$ be the customer that is being inserted. An ordered sequence of candidate insertion points, $L_{v_i} = \seq{v_1,\ldots,v_n}$ such that $v_j \in V - v_i$, is kept that lists customers in increasing geographic distance from $v_i$, that is $c_{ij} \forall j \in V - i$. The LNS repair operator tests $\mu$ positions from $L_{v_i}$ to find the cheapest insertion point. The repair operator tests both possible insertion points represented by $v_j \in L_{v_i}$, that is, it tests both the insertion cost of inserting $v_i$ immediately before and after $v_j$ in the route $R$ that contains $v_j$. 

For each site $s_i \in S$ we also maintain a counter $a_i, i \in S$ that denotes the age of the site. A site's age is incremented for each iteration that a site doesn't improve upon its currently best known solution (as defined by the solution's fitness, $f(\schd)$). Whenever a site improves upon its best known solution then the counter is reset, such that $a_i = 0$.

\picscl{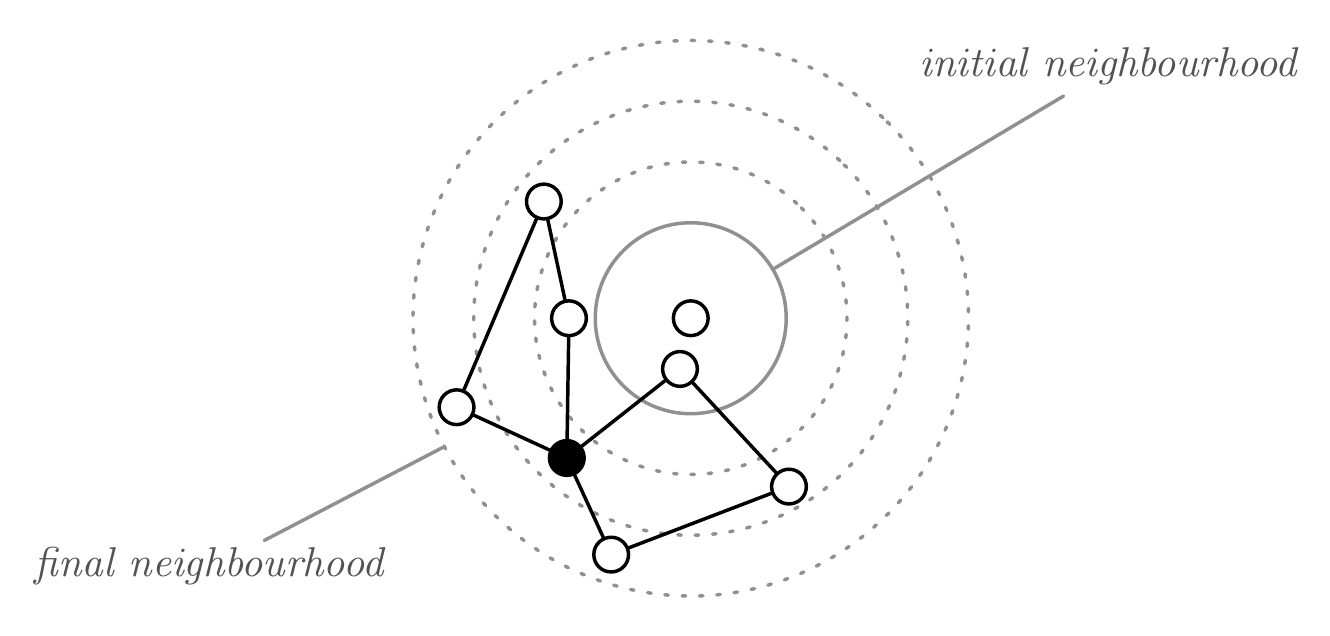}{The diagram shows a vertex being considered for reinsertion back into a solution. It starts by only considering insertion points that are geographically close to its position. The search area is widened over the course of the algorithm to take into consideration a larger set of insertion points.}{fig:neighbourhood}{0.85}

We then use the following formula to increase how much of $L_{v_i}$ is considered as the site ages.

\[
   \mu = \left|L_{v_i}\right| min\left(\frac{a_i}{k}, 1\right)
\]

Where $k$ is a constant that controls the rate at which the search area is expanded. 

As this process extends the number of insertion positions that are considered by the repair heuristic, this also serves to extend the neighbourhood of solutions surrounding a candidate solution $\schd$. In this way the algorithm also dynamically extends the size of a neighbourhood surrounding a site, $s_i \in S$, if $s_i$ becomes stuck in a local optimum.


\chapter{Results}
\label{chap:results}

In this chapter we provide a detailed breakdown of the results obtained by the Enhanced Bees Algorithm. The algorithm is tested against the well known set of test instances from Christofides, Mingozzi and Toth~\cite{CMT:1981}. We start in Section~\ref{sec:standardresults} by presenting the results obtained by running the algorithm in its two standard configurations: the first configuration is optimised to produce the best overall results, regardless of the runtime performance; the second configuration is optimised to produce the best results possible within a 60 second runtime threshold. We follow this in Section~\ref{sec:experiments} by providing contrasting results on the same problem instances instead using a standard Bees Algorithm and a LNS local search. Here we aim to demonstrate that the enhancements suggested in this thesis do in fact improve the solution quality. Finally, we end in Section~\ref{sec:comparison} by comparing and ranking how the Enhanced Bees Algorithm performs compared to the results in the literature. 

\section{Enhanced Bees Algorithm}
\label{sec:standardresults}

The results depicted in Figures~\ref{fig:standard_best},~\ref{fig:standard_best_blowup},~\ref{fig:standard_fast} and~\ref{fig:standard_fast_blowup} show the algorithm's performance on two configurations. The first configuration is optimised to produce the best overall results, that is, the minimum travel distance that meets all capacity and duration constraints. No consideration is made for the algorithm's runtime performance in this configuration. This configuration is denoted as \emph{Best} in the following diagrams and tables. The second configuration is optimised to produce the best results possible within a 60 second runtime window. This configuration is denoted as \emph{Fast} in the following diagrams and tables.


\picscl{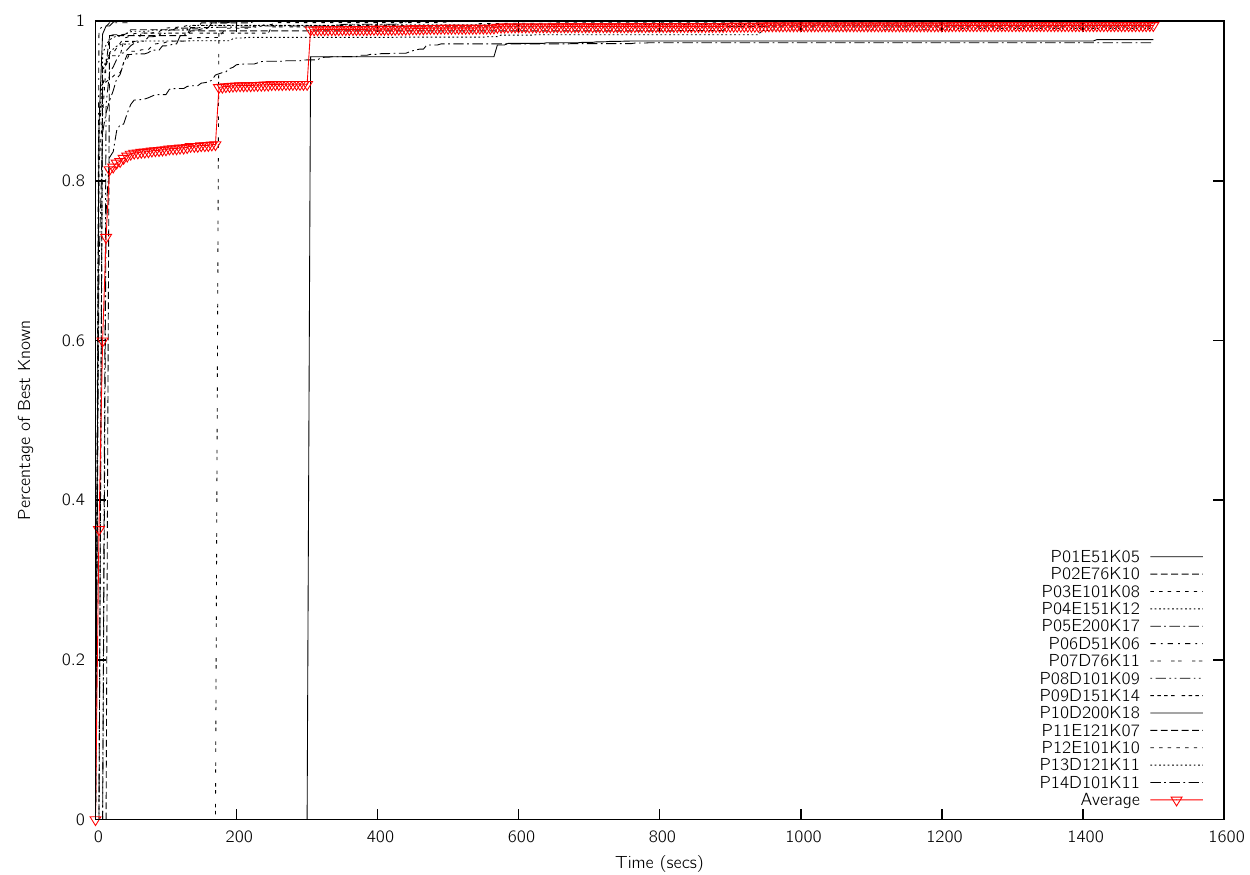}{\emph{Best}. Shown are the results obtained on the standard 14 Christofides, Mingozzi and Toth problem instances when the algorithm is optimised for producing the best overall results. The algorithm was left to run for 30 minutes. The lefthand axis shows the relative percentage compared to the best known result for the same problem instance. The bottom axis shows the elapsed runtime in seconds. Infeasible solutions are shown as being 0\% of the best known result. The average result obtained across all problem instances is shown in red.}{fig:standard_best}{1.1}

\picscl{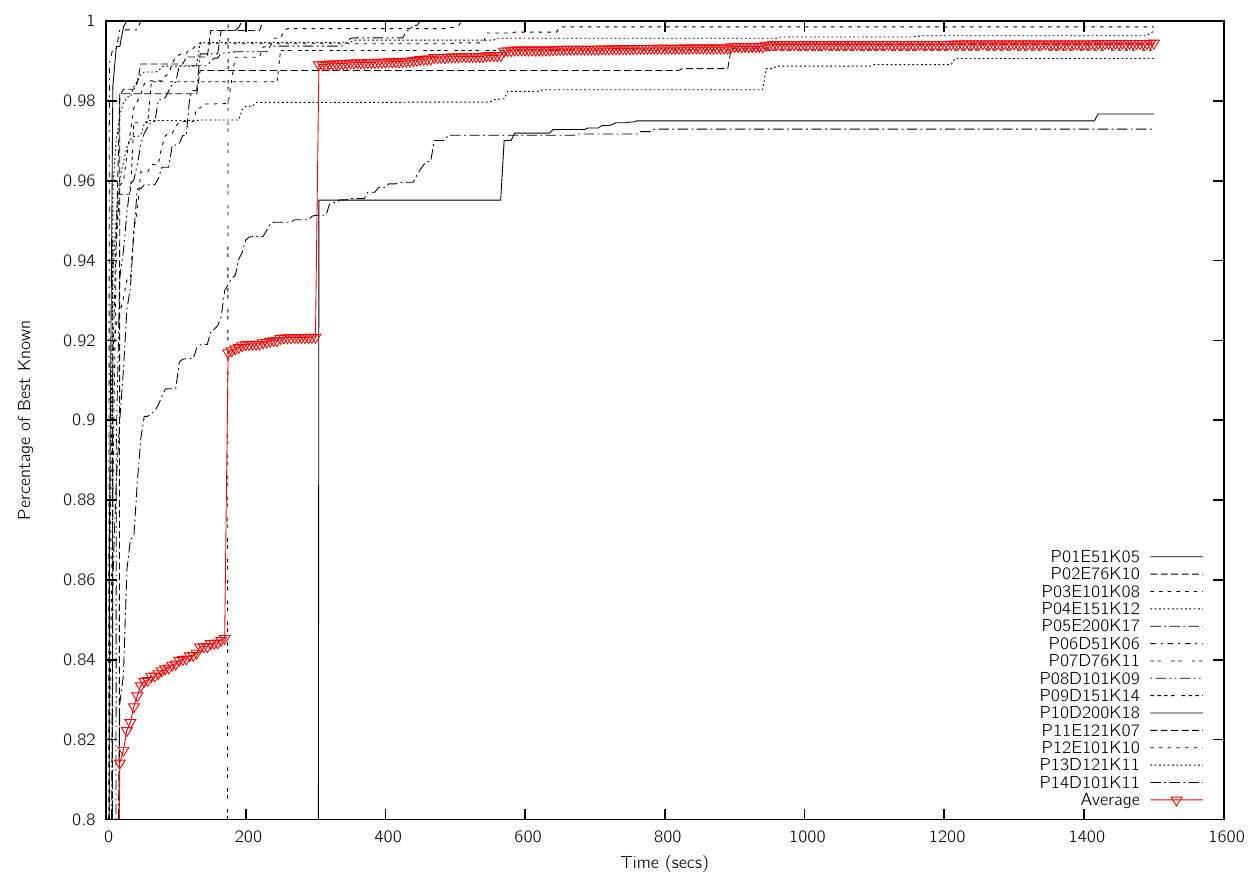}{\emph{Best---Blow up}. Shown are the same results as are depicted in Figure~\ref{fig:standard_best}, but with the area between 0.8 and 1.0 of the left axis blown up.}{fig:standard_best_blowup}{1.1}


\picscl{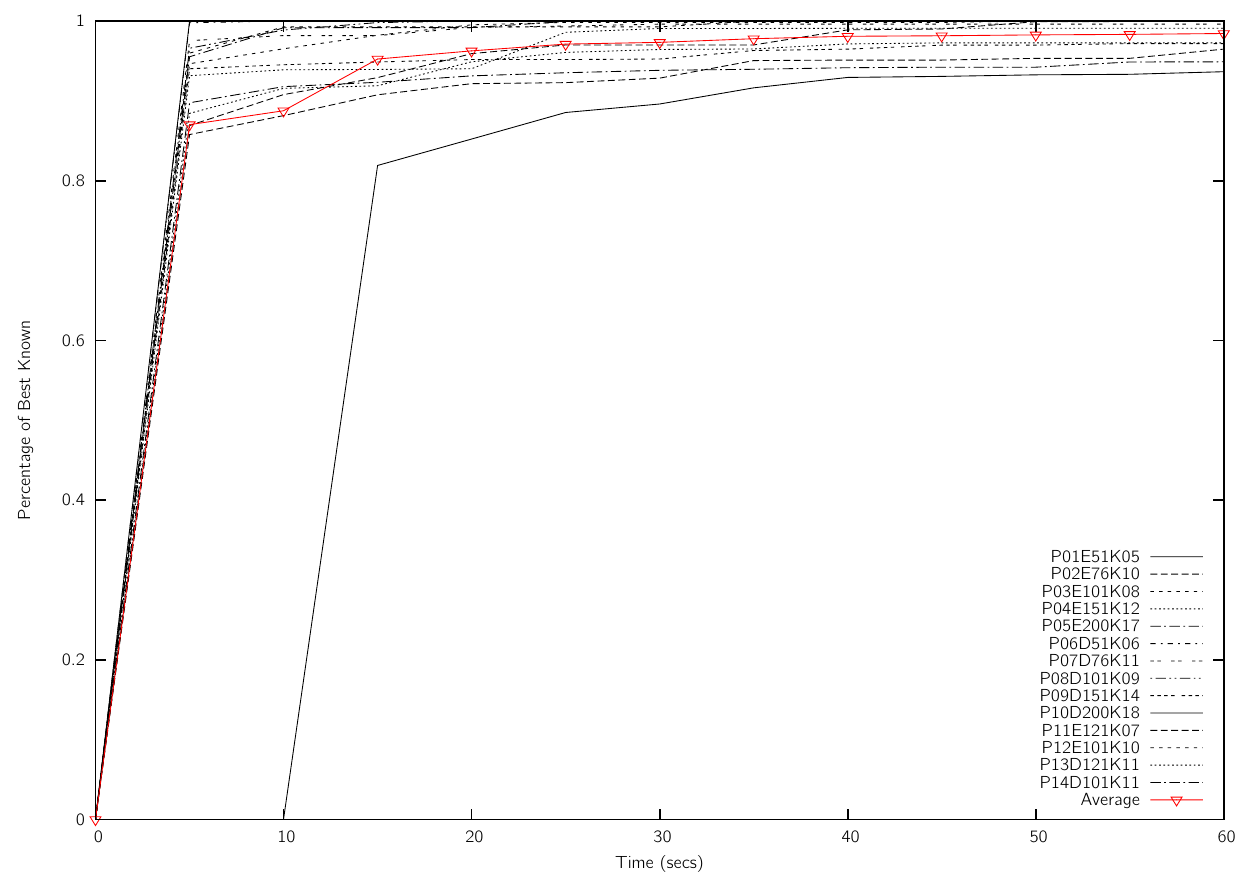}{\emph{Fast}. Shown are the results obtained on the standard 14 Christofides, Mingozzi and Toth problem instances when the algorithm is optimised for producing the best results within a 60 second runtime limit. The lefthand axis shows the relative percentage compared to the best known result for the same problem instance. The bottom axis shows the elapsed runtime in seconds. Infeasible solutions are shown as being 0\% of the best known result.}{fig:standard_fast}{1.1}

\picscl{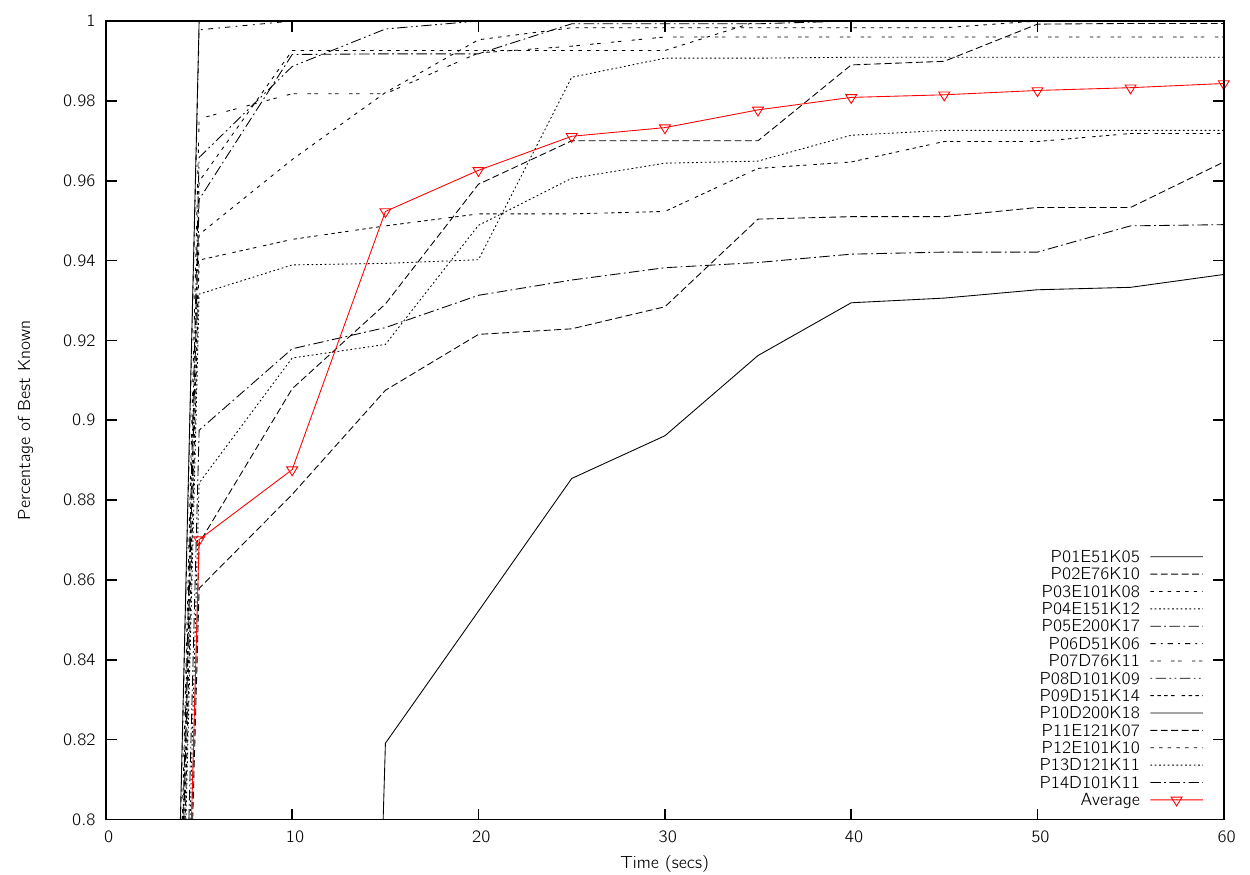}{\emph{Fast---Blow up}. Shown are the same results as are depicted in Figure~\ref{fig:standard_fast}, but with the area between 0.8 and 1.0 of the left axis blown up.}{fig:standard_fast_blowup}{1.1}

The results depicted in Figures~\ref{fig:standard_best},~\ref{fig:standard_best_blowup},~\ref{fig:standard_fast} and~\ref{fig:standard_fast_blowup} are for the standard 14 \VRP\ problem instances due to Christofides, Mingozzi and Toth~\cite{CMT:1981} and were all obtained on a MacBook Pro 2.8 GHz Intel Core 2 Duo. The best result for each problem instance was selected from 10 runs of the algorithm. 

In Figures~\ref{fig:standard_best} and~\ref{fig:standard_best_blowup} (i.e.~the \emph{Best} configuration) the algorithm was set to start with $|S| = 100$ (i.e.~100 sites), and reduce this number each 50 iterations ($\lambda = 50$) by 1\% until only $|S| = 3$. The number of promising solutions remembered by each site was set to $|M| = 5$. The algorithm was left to run for 30 minutes on each problem before being terminated. Infeasible solutions (i.e.~solutions over their capacity or duration constraints) were allowed to be traversed through, but were scored as being 0\% of the best known solution.

Conversely, in Figures~\ref{fig:standard_fast} and~\ref{fig:standard_fast_blowup} (i.e.~the \emph{Fast} configuration) the algorithm was set to start with $|S| = 25$ (i.e.~25 sites), and reduce this number each iteration ($\lambda = 1$) by 1\% until only $|S| = 1$. The number of promising solutions remembered by each site was set to $|M| = 5$. The algorithm was left to run for 60 seconds on each problem before being terminated. The LNS improvement heuristic was set to destroy between 0\% and 80\% (with a mean of 40\%) of the solution at each step. The repair operator initially only considers the first 3 closest customers as reinsertion points, but it increases this to 50\% of the closest customers as the site ages. Infeasible solutions (i.e.~solutions over duration or service time constraints) were allowed to be traversed through, but were scored as being 0\% of the best known solution.

Table~\ref{tab:standard_results} provides a tabular summary of the results covered in this section.

\ctable[
   caption={\emph{Enhanced Bees Algorithm}. Given are the results obtained by running the Enhanced Bees Algorithm using two different configurations on the standard 14 Christofides, Mingozzi and Toth problem problem instances. The \emph{fast} configuration is optimised to produce results within a 60 second runtime limit. Conversely, the \emph{best} configuration is optimised for producing the best results within the much longer period of 30 minutes.}, 
   label=tab:standard_results, 
   pos=htb, 
   framesep=10pt]
{lr>{\itshape}rr>{\itshape}rr}
{
   \tnote[1]{Run for 60 seconds.}
   \tnote[2]{Run for 30 minutes.}
   \tnote[3]{As reported by Gendreau, Laporte, and Potvin in~\cite{GLP:1999}.}
}{
\FL
   Instance
   & \multicolumn{4}{c}{Results}
   & Best Known\tmark[3]
\NN
   & \multicolumn{2}{c}{Fast\tmark[1]} & \multicolumn{2}{c}{Best\tmark[2]}
\ML
   P01E51K05 & 524.61   & 100.00\%  & 524.61    & 100.00\%  & 524.61 \\
   P02E76K10 & 835.77   & 99.94\%   & 835.26    & 100.00\%  & 835.26  \\
   P03E101K08 & 826.14  & 100.00\%  & 826.14    & 100.00\%  & 826.14  \\
   P04E151K12 & 1057.40 & 97.26\%   & 1036.12   & 99.26\%   & 1028.42  \\
   P05E200K17 & 1360.85 & 94.90\%   & 1327.48   & 97.29\%   & 1291.45  \\
   P06D51K06 & 555.43   & 100.00\%  & 555.43    & 100.00\%  & 555.43  \\
   P07D76K11 & 913.37   & 99.60\%   & 909.68    & 100.00\%   & 909.68  \\
   P08D101K09 & 865.94  & 100.00\%  & 865.94    & 100.00\%  & 865.94  \\
   P09D151K14 & 1196.32 & 97.18\%   & 1169.24   & 99.43\%   & 1162.55  \\
   P10D200K18 & 1490.49 & 93.65\%   & 1428.54   & 97.71\%   & 1395.85  \\
   P11E121K07 & 1080.20 & 96.47\%   & 1048.24   & 99.42\%   & 1042.11  \\
   P12E101K10 & 819.56  & 100.00\%  & 819.56    & 100.00\%  & 819.56  \\
   P13D121K11 & 1555.30 & 99.09\%   & 1545.19   & 99.74\%   & 1541.14  \\
   P14D101K11 & 866.37  & 100.00\%  & 866.37    & 100.00\%  & 866.37
\ML
   Average    &         & 98.44\%  &            & 99.49\%   &
\LL
}

\section{Experiments}
\label{sec:experiments}

In this section we review the results obtained by implementing the standard Bees Algorithm and a LNS local search. The aim of these experiments is to prove that the algorithmic enhancements suggested in this thesis do in fact produce better results than would have been obtained if we had used a standard Bees Algorithm. We also demonstrate that the combination of the Bees Algorithm with the LNS local search produces better results than if either algorithm were used separately.

\subsection{Bees Algorithm versus Enhanced Bees Algorithm}
\label{subsec:bavebs}

We start in Figures~\ref{fig:truebees_nolns} and~\ref{fig:truebees_nolns_blowup} by showing the results obtained by using the standard Bees Algorithm as described by Pham et al.~in~\cite{PGKORZ:2005}. The same problem instances as Section~\ref{sec:standardresults} are used, so that the results can be compared directly.

\picscl{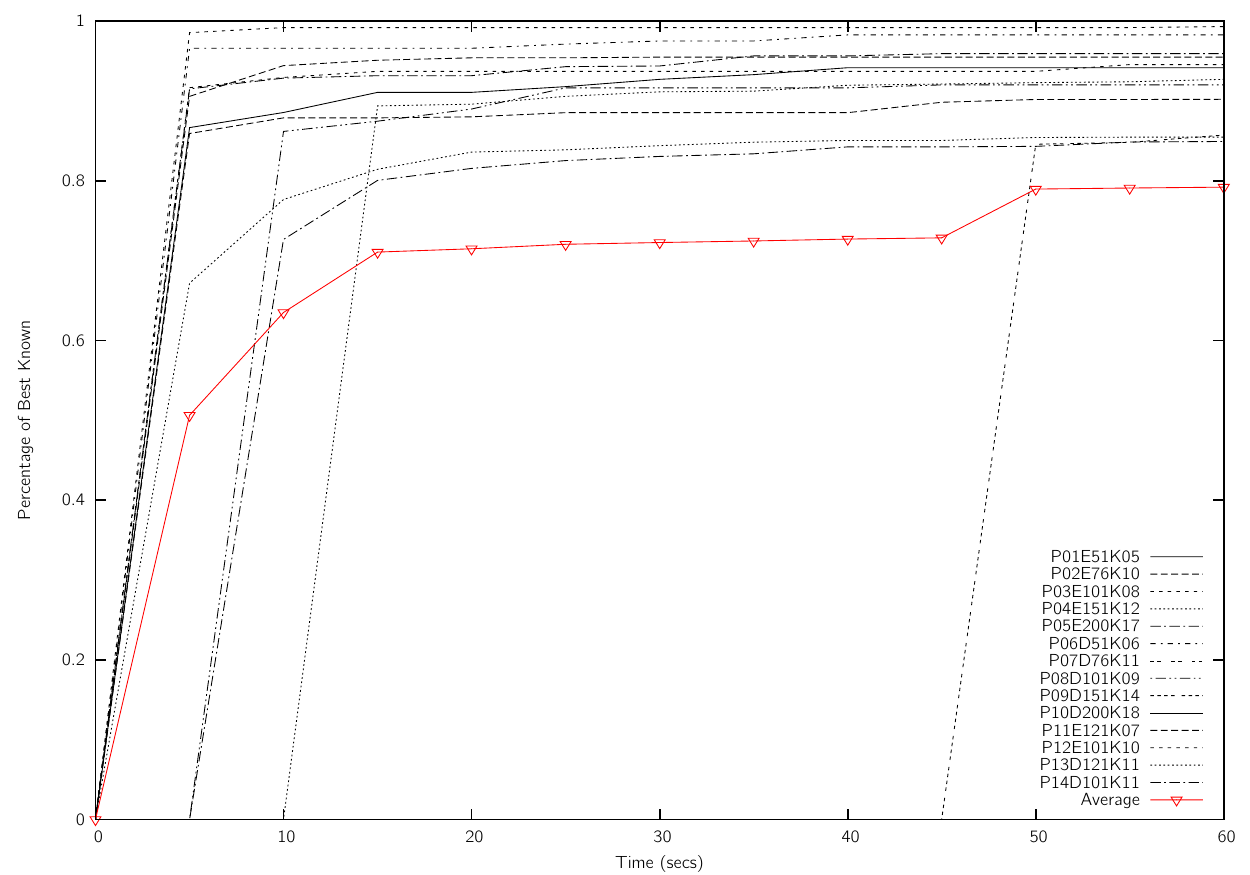}{\emph{Standard Bees Algorithm}. Shown are the results obtained on the standard 14 Christofides, Mingozzi and Toth instances when the algorithm is optimised for producing the best results within a 60 second runtime limit. The lefthand axis shows the relative percentage compared to the best known result for the same problem instance. The bottom axis shows the elapsed runtime in seconds. Infeasible solutions are shown as being 0\% of the best known result. The average result obtained across all problem instances is shown in red.}{fig:truebees_nolns}{1.1}

\picscl{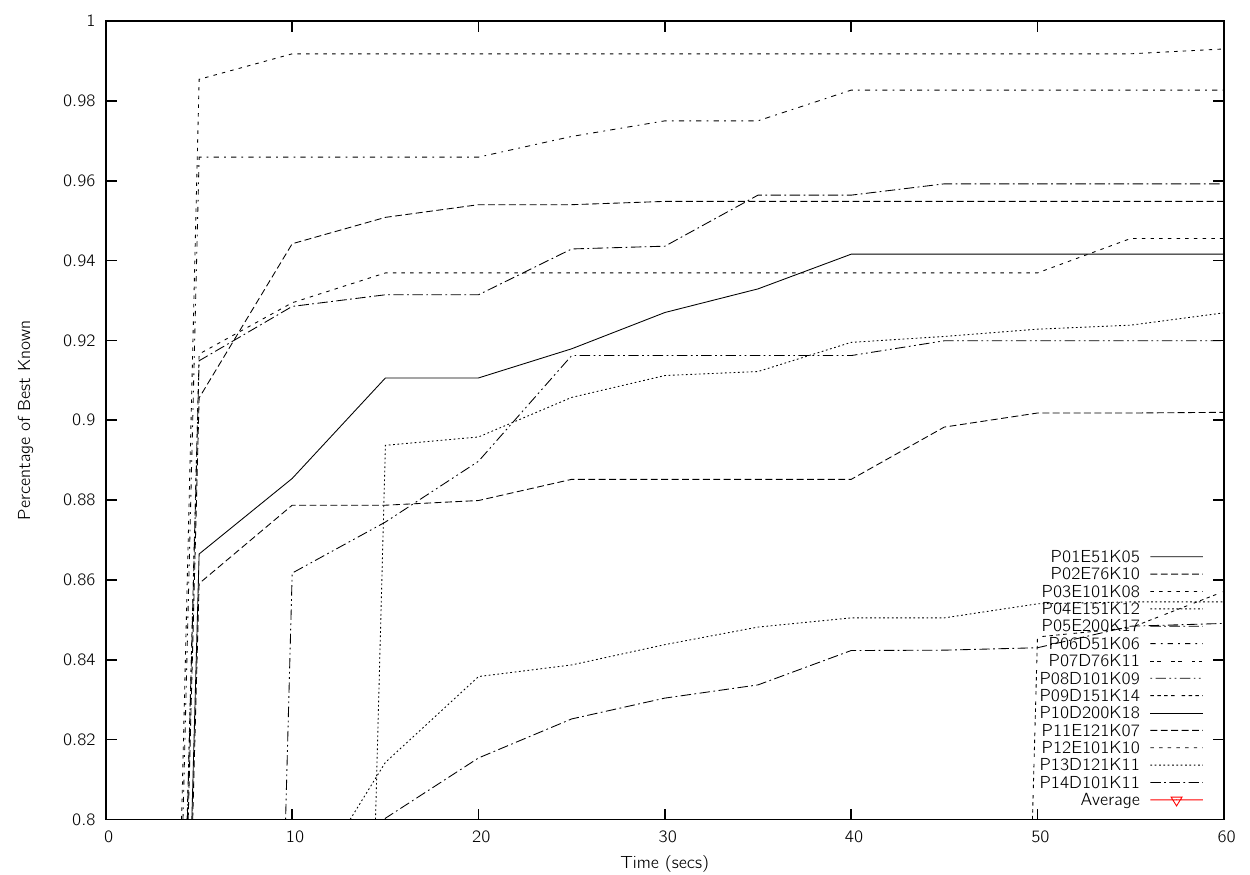}{\emph{Standard Bees Algorithm---Blow up}. Shown are the same results as are depicted in Figure~\ref{fig:truebees_nolns}, but with the section between 0.8 and 1.0 of the left axis blown up. Note that the average is not shown on the blow up because it is below 80\% percent.}{fig:truebees_nolns_blowup}{1.1}

The results depicted in Figures~\ref{fig:truebees_nolns} and~\ref{fig:truebees_nolns_blowup} were obtained on a MacBook Pro 2.8 GHz Intel Core 2 Duo. The best result for each instance was selected from 10 runs of the algorithm. The algorithm was configured with the following parameters. It used 25 sites, and selected the best 6 sites as being elite. Each elite site had 3 bees recruited for the search. Another 6 sites were selected as being non-elite, and had 2 bees recruited for the search. The bees from the remaining 13 sites were left to search randomly. The algorithm was left to run for 60 seconds on each problem before being terminated. Infeasible solutions (i.e.~solutions over their capacity or duration constraints) were allowed to be traversed through, but were scored as being 0\% of the best known solution. A $\lambda$-interchange (with $\lambda = 2$) improvement heuristic was used for the improvement phase of each bee (see Chapter~\ref{chap:background} for an overview on how this heuristic works).

\subsection{Large Neighbourhood Search}
\label{subsec:largeneighbourhoodsearch}

Next we show in Figures~\ref{fig:lns} and~\ref{fig:lns_blowup} the results obtained by using a standalone LNS search embedded within a hill climb meta-heuristic. The LNS search used in this section is the same one that is employed by the Enhanced Bees Algorithm. It should be noted that there are more sophisticated LNS algorithms available than the comparatively simple one used here. And that these would most probably produce better results than the LNS results presented here. However, we believe one of the attractive features of the Enhanced Bees Algorithm is that it uses a fairly simple local search procedure. Moreover, our aim in this experiment was to demonstrate that the limitations of our simple local method are offset by it being embedded within a Bees Algorithm.

\picscl{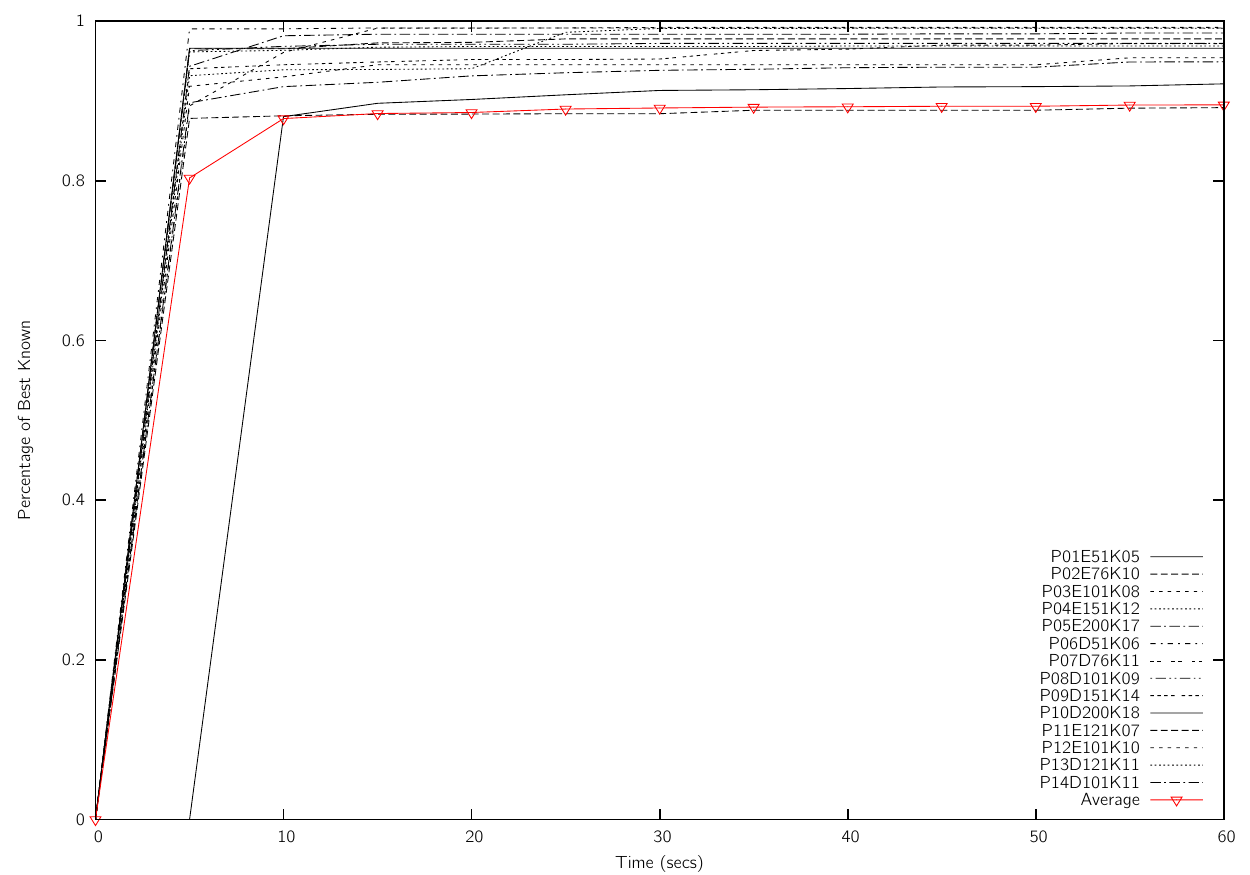}{\emph{LNS algorithm}. Shown are the results obtained on the standard 14 Christofides, Mingozzi and Toth instances when the algorithm is optimised for producing the best results within a 60 second runtime limit. The lefthand axis shows the relative percentage compared to the best known result for the same problem instance. The bottom axis shows the elapsed runtime in seconds. Infeasible solutions are shown as being 0\% of the best known result. The average result obtained across all problem instances is shown in red.}{fig:lns}{1.1}

\picscl{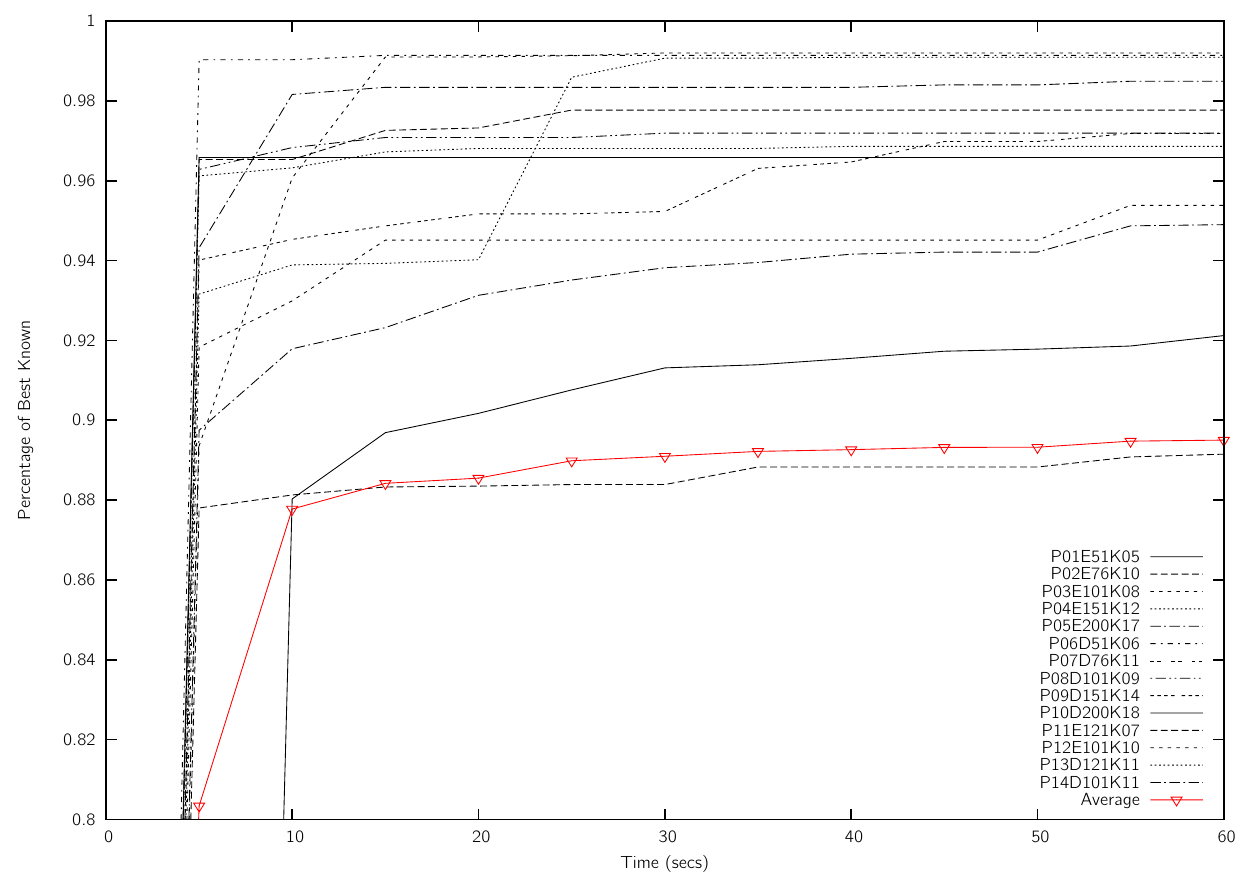}{\emph{LNS algorithm---Blow up}. Shown are the same results as are depicted in Figure~\ref{fig:lns}, but with the area between 0.8 and 1.0 of the left axis blown up.}{fig:lns_blowup}{1.1}

The results depicted in Figures~\ref{fig:lns} and~\ref{fig:lns_blowup} were obtained on a MacBook Pro 2.8 GHz Intel Core 2 Duo. The best result for each problem instance was selected from 10 runs of the algorithm. The algorithm was initialised to a starting position generated by a simple insertion heuristic. The LNS heuristic was set to destroy between 0\% and 80\% (with a mean of 40\%) of the solution at each step. The repair enumerated all customers when deciding the best reinsertion point. Infeasible solutions (i.e.~solutions over their capacity or duration constraints) were allowed to be traversed through, but were scored as being 0\% of the best known solution.

\subsection{Summary}

Table~\ref{tab:results_summary} provides a summary of the results obtained by the Bees Algorithm and LNS experiments, alongside the results obtained by our Enhanced Bees Algorithm. The Bees Algorithm is the worst of the three. This is not surprising given that the Bees Algorithm was devised to solve continuous problems, rather than discrete problems. Where the Bees Algorithm has been used for discrete problems in the literature it has been adapted to incorporate more sophisticated local search techniques, much as the Enhanced Bees Algorithm has been here. Also of note is that two of the problem instances didn't produce feasible solutions at all within the 60 second runtime threshold. We believe that given a longer running time the algorithm would most probably have found a feasible solution. However, one of the objectives of the Enhanced Bees Algorithm is to produce robust results reliably; in this count the standard Bees Algorithm is not competitive. 

The LNS improvement heuristic produced much stronger results. This shows that the LNS improvement plays an important part in the results obtained by the Enhanced Bees Algorithm. The LNS heuristic is a fairly new heuristic (in the \VRP\ research at least); nevertheless it has produced some of the most competitive results. This is borne out by the results obtained here and in Section~\ref{sec:standardresults}. However, the LNS local search did fail to find a feasible solution for one of the problem instances. Again this is probably due to the limited runtime permitted. If this problem instance is removed from the results, then the LNS search's average result becomes 96.39\%, getting us much closer to the results obtained by the Enhanced Bees Algorithm.

\ctable[
   caption={\emph{Experiments}. Given are results that contrast the standard Bees Algorithm, a LNS local search, and the Enhanced Bees Algorithm with one another. Each algorithm was run for 60 seconds and used the standard 14 Christofides, Mingozzi and Toth problem instances.},
   label=tab:results_summary,
   pos=h,
   framesep=10pt]
{lr>{\itshape}rr>{\itshape}rr>{\itshape}r}
{
   \tnote[1]{Run for 60 seconds and configured as described in Section~\ref{subsec:bavebs}.}
   \tnote[2]{Run for 60 seconds and configured as described in Section~\ref{subsec:largeneighbourhoodsearch}.}
   \tnote[3]{Run for 60 seconds and configured as described in Section~\ref{sec:standardresults}.}
}{
\FL
   Instance
   & \multicolumn{2}{c}{Bees Algorithm\tmark[1]}
   & \multicolumn{2}{c}{LNS\tmark[2]}
   & \multicolumn{2}{c}{Enhanced Bees\tmark[3]}
\ML
   P01E51K05 & 557.17   & 94.16\%   & 543.16    & 96.58\%    & 524.61   & 100.00\%  \\
   P02E76K10 & 925.97   & 90.20\%   & 854.35    & 97.77\%    & 835.77   & 99.94\%   \\
   P03E101K08 & 873.75  & 94.55\%   & 866.13    & 95.38\%    & 826.14   & 100.00\%  \\
   P04E151K12 & 1203.47 & 85.45\%   & 1061.75   & 96.86\%    & 1057.40  & 97.26\%   \\
   P05E200K17 & 1520.89 & 84.91\%   & 1360.85   & 94.90\%    & 1360.85  & 94.90\%   \\
   P06D51K06 & 565.21   & 98.27\%   & 560.24    & 99.14\%    & 555.43   & 100.00\%  \\
   P07D76K11 & 1153.57  & 00.00\%   & 1215.12   & 00.00\%    & 913.37   & 99.60\%   \\
   P08D101K09 & 941.34  & 91.99\%   & 890.93    & 97.19\%    & 865.94   & 100.00\%  \\
   P09D151K14 & 1356.43 & 85.71\%   & 1196.31   & 97.18\%    & 1196.32  & 97.18\%   \\
   P10D200K18 & 2153.48 & 00.00\%   & 1515.32   & 92.12\%    & 1490.49  & 93.65\%   \\
   P11E121K07 & 1091.42 & 95.48\%   & 1168.91   & 89.15\%    & 1080.20  & 96.47\%   \\
   P12E101K10 & 825.38  & 99.30\%   & 826.14    & 99.20\%    & 819.56   & 100.00\%  \\
   P13D121K11 & 1662.70 & 92.69\%   & 1555.29   & 99.09\%    & 1555.30  & 99.09\%   \\
   P14D101K11 & 903.25  & 95.92\%   & 879.69    & 98.49\%    & 866.37   & 100.00\%
\ML
   Average    &         & 79.18\%  &            & 89.50\%   &           & 98.44\%
\LL
}

\section{Comparison}
\label{sec:comparison}

Lastly, in Table~\ref{tab:resultscomparison} and Figure~\ref{fig:resultscomparison} we provide a comparison of the Enhanced Bees Algorithm along with other well known results from the literature. As can be seen from Table~\ref{tab:resultscomparison} and Figure~\ref{fig:resultscomparison} some of the best results known are due to Taillard's Tabu Search heuristic. He reaches 12 of the best known solutions from the set of 14 problems. The Enhanced Bees Algorithm, by comparison, finds 8 of the 14 best known solutions. However, the Enhanced Bees Algorithm is still very competitive. The runtime duration required to find a best known solution is smaller than many of the other meta-heuristics (although a direct comparison is hard to make as many of the reported results were run on significantly older hardware). Additionally, the solutions produced by the Enhanced Bees Algorithm are within 0.5\% of the best known solutions on average, meaning that the algorithm is very competitive with the best meta-heuristics available for the \VRP.

\ctable[
   caption={\emph{Results Comparison}. Given is a comparison of the Enhanced Bees Algorithm alongside other well known results from the literature. The results are for the standard 14 Christofides, Mingozzi and Toth problem instances commonly used in the \VRP\ literature. Running times are given in parenthesis where known.},
   label=tab:resultscomparison, 
   framesep=10pt, 
   sideways, 
   doinside=\small]
{lrrrrrrrrr}
{
   \tnote[1]{Clark Write's Savings (Parallel) algorithm. Implemented by Laporte and Semet~\cite{Laporte:1999}.}
   \tnote[2]{Sweep Algorithm due to Gillett and Miller~\cite{GM:1974}. Implemented by Christofides, Mingozzi and Toth~\cite{CMT:1981}. Reported in~\cite{Laporte:1999}.}
   \tnote[3]{Generalised Assignment due to Fisher and Jaikumar~\cite{FJ:1981}. Reported in~\cite{Laporte:1999}.}
   \tnote[4]{3-Opt local search applied after Clark Write's Savings (Parallel) algorithm. First improvement taken. Implemented by Laporte and Semet~\cite{Laporte:1999}.}
   \tnote[5]{Simulated Annealing due to Osman~\cite{Osman:1993}. Runtime duration is given in parentheses and is reported in seconds on a VAX 86000.}
   \tnote[6]{Tabu Search due to Taillard~\cite{Taillard:1993}. Runtime duration is given in parentheses and is reported in seconds on a Sillicon Graphics Workstation, 36Mhz.}
   \tnote[7]{Ant Colony Optimisation  Bullnheimer, Hartl, Strauss~\cite{BHS:1999B}. Bullnheimer et al.~provided two papers on Ant Colony Optimisation for \VRP, the better of the two is used. Runtime duration is given in parentheses and is reported in seconds on a Pentium 100.}
   \tnote[8]{Enhanced Bees Algorithm. Results are shown from the \emph{best} configuration in Section~\ref{sec:standardresults}. Runtime duration is given in parentheses and is reported in seconds on a MacBook Pro 2.8 GHz Intel Core 2 Duo.}
   \tnote[9]{Best known results as reported by Gendreau, Laporte, and Potvin in~\cite{GLP:1999}.}
}{
\FL
   Instance
   & CW\tmark[1]
   & Sweep\tmark[2]
   & GenAsgn\tmark[3]
   & 3-Opt\tmark[4]
   & SA\tmark[5]
   & TS\tmark[6] 
   & ACO\tmark[7] 
   & EBA\tmark[8]
   & Best Known\tmark[9]
\ML
   P01E51K05   & 584.64    & 532    & 524    & 578.56    & 528 (167)             & \textbf{524.61} (360)    & \textbf{524.61} (6)   & \textbf{524.61} (5)   & 524.61  \\
   P02E76K10   & 900.26    & 874    & 857    & 888.04    & 838.62 (6434)         & \textbf{835.26} (3228)   & 844.31 (78)           & \textbf{835.26} (225) & 835.26  \\
   P03E101K08  & 886.83    & 851    & 833    & 878.70    & 829.18 (9334)         & \textbf{826.14} (1104)   & 832.32 (228)          & \textbf{826.14} (50)  & 826.14  \\
   P04E151K12  & 1133.43   & 1079   & 1014   & 1128.24   & 1058 (5012)           & \textbf{1028.42} (3528)  & 1061.55 (1104)        & 1036.12 (1215)        & 1028.42 \\
   P05E200K17  & 1395.74   & 1389   & 1420   & 1386.84   & 1378 (1291)           & 1298.79 (5454)           & 1343.46 (5256)        & 1327.48 (785)         & 1291.45 \\
   P06D51K06   & 618.40    & 560    & 560    & 616.66    & \textbf{555.43} (3410)& \textbf{555.43} (810)    & 560.24 (6)            & \textbf{555.43} (5)   & 555.43  \\
   P07D76K11   & 975.46    & 933    & 916    & 974.79    & 909.68 (626)          & \textbf{909.68} (3276)   & 916.21 (102)          & \textbf{909.68} (175) & 909.68  \\
   P08D101K09  & 973.94    & 888    & 885    & 968.73    & 866.75 (957)          & \textbf{865.94} (1536)   & 866.74 (288)          & \textbf{865.94} (20)  & 865.94  \\
   P09D151K14  & 1287.64   & 1230   & 1230   & 1284.64   & 1164.12 (84301)       & \textbf{1162.55} (4260)  & 1195.99 (1650)        & 1169.24 (1610)        & 1162.55 \\
   P10D200K18  & 1538.66   & 1518   & 1518   & 1538.66   & 1417.85 (5708)        & 1397.94 (5988)           & 1451.65 (4908)        & 1428.54 (1540)        & 1395.85 \\
   P11E121K07  & 1071.07   & 1266   & -      & 1049.43   & 1176 (315)            & \textbf{1042.11} (1332)  & 1065.21 (552)         & 1048.24 (960)         & 1042.11 \\
   P12E101K10  & 833.51    & 937    & 824    & 824.42    & 826 (632)             & \textbf{819.56} (960)    & \textbf{819.56} (300) & \textbf{819.56} (35)  & 819.56  \\
   P13D121K11  & 1596.72   & 1776   & -      & 1587.93   & 1545.98 (7622)        & \textbf{1541.14} (3552)  & 1559.92 (660)         & 1545.19 (1500)        & 1541.14 \\
   P14D101K11  & 875.75    & 949    & 876    & 868.50    & 890 (305)             & \textbf{866.37} (3942)   & 867.07 (348)          & \textbf{866.37} (40)  & 866.37  \\
\LL
}

\picscl{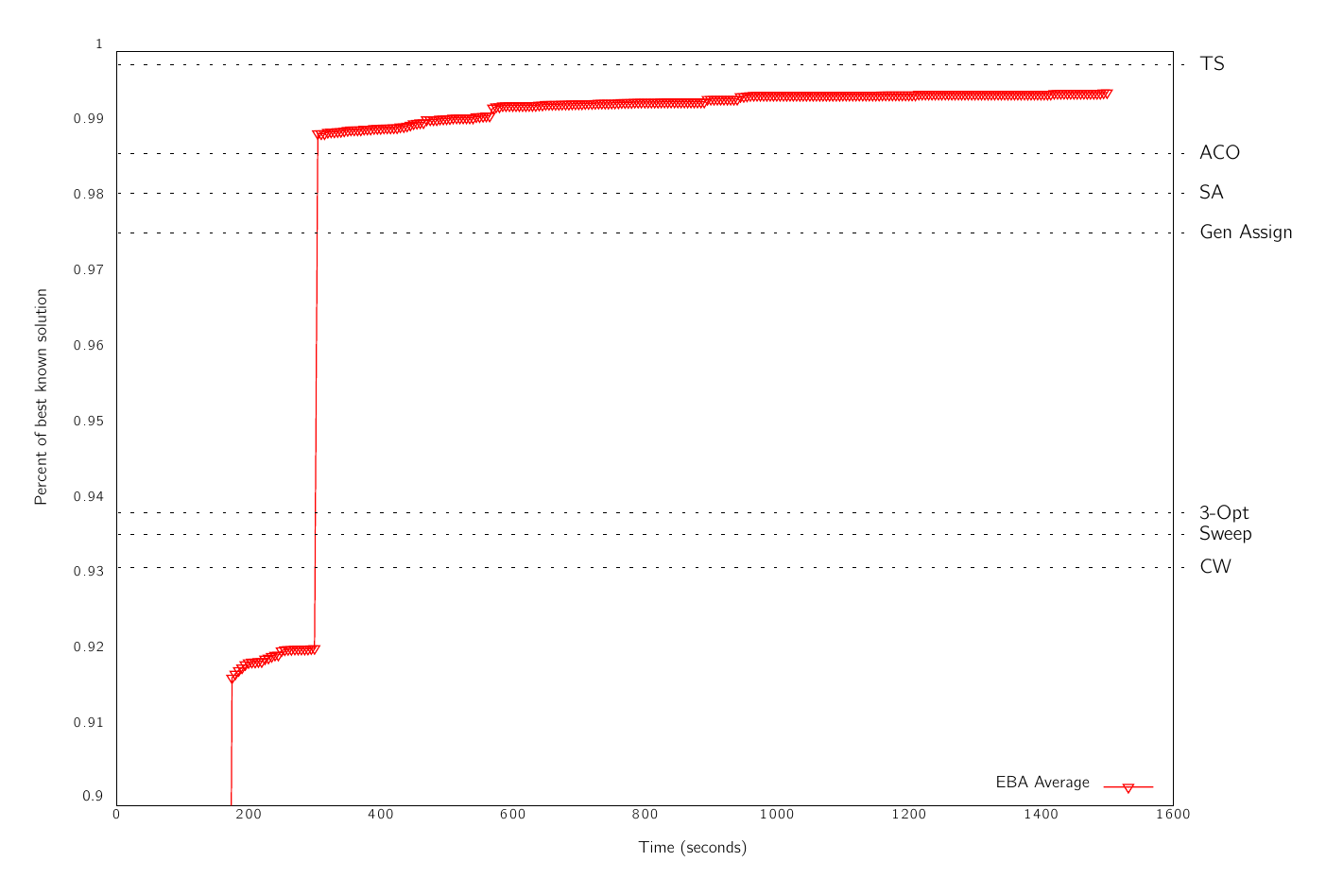}{\emph{Results Comparison}. Shown are the average results across all problem instances obtained by the Enhanced Bees Algorithm (shown in red) mapped against the results listed in Table~\ref{tab:resultscomparison}.}{fig:resultscomparison}{1.0}


\chapter{Conclusion}
\label{chap:conclusion}

In this thesis we have described a new meta-heuristic for the \VRP\ called the Enhanced Bees Algorithm. The results obtained are competitive with the best meta-heuristics available for the Vehicle Routing Problem. Additionally, the algorithm has good runtime performance, producing results within 2\% of the optimal solution within 60 seconds.

We took the Bees Algorithm as our starting point. The Bees Algorithm was originally developed for solving continuous optimisation problems, so part of the work undertaken for this thesis was to adapt it for use on the \VRP\ (although, it could be argued that the Enhanced Bees Algorithm is more `inspired by' than `adapted from'). The approach developed for the Enhanced Bees Algorithm could equally be applied to other combinatorial optimisation problems, such as the Traveling Salesman Problem, the Job Shop Scheduling Problem, or the Cutting Stock Problem.

We showed empirically that the quality of the solutions obtained by the Enhanced Bees Algorithm are competitive with the best modern meta-heuristics available for the \VRP. Additionally, we showed that the algorithm has good runtime performance, producing results within 2\% of the optimal solution within 60 seconds. The runtime performance of the algorithm makes it suitable for use within real world dispatch scenarios, where often the dispatch process is fluid and hence it is impractical for the optimisation to take minutes (or hours). In these environments, it is acceptable to trade a fraction of a percent off the solution quality for quicker runtime performance.

We also gave results that demonstrated that the algorithmic enhancements suggested in this thesis did in fact produce better results than would have been obtained by using a standard Bees Algorithm. We also demonstrated that the combination of the Bees Algorithm with a LNS local search heuristic produced better results than if the algorithms had been used separately.

Additionally, we provided a comprehensive survey of the \VRP\ literature. In this we provided a short history of the results that have been foundational to \VRP\ research, as well as providing in depth material and descriptions for some of the classic algorithms developed for the \VRP. 

There are a number of areas where the research undertaken in this thesis could be continued. These include:

\begin{itemize}

   \item \emph{Introduce a mating process to generate new sites}. An interesting extension to the Enhanced Bees Algorithm would be to incorporate a crossover operation for generating new sites. In our version of the Enhanced Bees Algorithm unpromising sites are simply culled off. An alternative approach, borrowing a concept from Genetic Algorithms and Tabu Search (in this case, Taillard's Adaptive Memory), would be to replace the site with a recombination of two other successful sites. The advantage to this approach is that it would open a new area of the search space to exploration; an area that should be promising as it combines components from two already successful sites. The crossover process could be as simple as using an existing \VRP\ crossover operator, such as $OX$, on the two fittest solutions from each site. Although much more sophisticated crossover operations are imaginable.

   \item \emph{Extend to other combinatorial problems}. It should be possible to follow a similar approach that we have taken in adapting the Bees Algorithm to the \VRP\ and apply this to other combinatorial problems. The Bees Algorithm is especially strong at providing robust solutions where the search space contains many local optima. We imagine that there are many other combinatorial problems where this would be an advantage. An obvious starting point would be to apply it to the Job Shop Scheduling Problem, which shares many characteristics with the \VRP.

   \item \emph{Include more real world constraints}. Although the literature is filled with variations of the \VRP\ that add additional constraints (i.e.~\VRPTW, \PDP, etc), the focus has been on tackling the computationally hard constraints, such as time windows, multiple deploys, etc. An area that is not often addressed in the literature is how to deal with soft constraints, such as, on the day disruptions, differing skill sets across a fleet, or factoring in a dispatcher's assignment preferences. Unfortunately, these constraints are a barrier to vehicle route optimisation being adopted in many logistics companies. 
   
   An interesting line of research would be to extend the optimisation methods developed for the \VRP\ to include feedback from a dispatcher---we envision an interactive process where the dispatcher can feedback into the optimisation process the soft constraints that are not modelled within the algorithm. It would also be very interesting to see if the supervised learning methods from artificial intelligence, such as the Naive Bayes classifier, could be incorporated into the optimisation process.
      
\end{itemize}



\bibliographystyle{plain}
\bibliography{bibliography}

\begin{thebibliography}{10}

\bibitem{beeimage}
Bee image.
\newblock \texttt{http://etc.usf.edu/clipart/27800/27890/bee\_27890.htm}.

\bibitem{beesalg}
The bees algorithm website.
\newblock \texttt{http://www.bees-algorithm.com}.

\bibitem{biodidac:website}
Biodidac, a bank of digital resources for teaching biology.
\newblock \texttt{http://biodidac.bio.uottawa.ca/}.

\bibitem{AG:1991}
K.~Altinkemer and B.~Gavish.
\newblock Parallel savings based heuristic for the delivery problem.
\newblock {\em Operations Research}, 39:456--469, 1991.

\bibitem{balinski:64}
M.L. Balinski and R.E. Quandt.
\newblock On an integer program for a delivery problem.
\newblock {\em Operations Research}, 12(2):300--304, 1964.

\bibitem{BM:2003}
Jean Berger and Mohamed Barkaoui.
\newblock A hybrid genetic algorithm for the capacitated vehicle routing
  problem.
\newblock In {\em Genetic and Evolutionary Computation---GECCO 2003}, volume
  2723 of {\em Lecture Notes in Computer Science}, pages 198--198. 2003.

\bibitem{BW:1993}
J.L. Blanton and R.L. Wainwright.
\newblock Multiple vehicle routing with time and capacity constraints using
  genetic algorithms.
\newblock {\em Proceedings of the 5th International Conference on Genetic
  Algorithms}, pages 451--459, 1993.

\bibitem{BHS:1999A}
B.~Bullnheimer, R.~F. Hartl, and C.~Strauss.
\newblock Applying the ant system to the vehicle routing problem.
\newblock {\em Meta-heuristics: Advances and trends in local search paradigms
  for optimization}, pages 285--296, 1999.

\bibitem{BHS:1999B}
B.~Bullnheimer, R.~F. Hartl, and C.~Strauss.
\newblock An improved ant system algorithm for the vehicle routing problem.
\newblock {\em Annals of Operations Research}, 89:319–328, 1999.

\bibitem{CE:1969}
N.~Christofides and S.~Eilon.
\newblock An algorithm for the vehicle dispatching problem.
\newblock {\em ORQ}, 20:309--318, 1969.

\bibitem{CMT:1981}
N.~Christofides, A.~Mingozzi, and P.~Toth.
\newblock Exact algorithms for the vehicle routing problem, based on spanning
  tree and shortest path relaxations.
\newblock {\em Mathematical Programming}, 20:255--282, 1981a.

\bibitem{clark:1964}
G.~Clark and J.W. Wright.
\newblock Scheduling of vehicles from a central depot to a number of delivery
  points.
\newblock {\em Operations Research}, 12:568--581, 1964.

\bibitem{Dantzig:1954}
Dantzig, Fulkerson, and Johnson.
\newblock Solution of a large-scale traveling salesman problem.
\newblock {\em Operations Research}, (2):393–410, 1954.

\bibitem{Dantzig:1959}
G.~B. Dantzig and J.~H. Ramser.
\newblock The truck dispatching problem.
\newblock {\em Management Science}, 6(1):80--91, Oct, 1959.

\bibitem{DMDJSM:1992}
Martin Desrochers, Jacques Desrosiers, and Marius Solomon.
\newblock A new optimization algorithm for the vehicle routing problem with
  time windows.
\newblock {\em Operations Research}, 40:342--354, March 1992.

\bibitem{EWC:1971}
S.~Eilon, C.D. Watson-Gandy, and N.~Christofides.
\newblock {\em Distribution management: Mathematical modelling and practical
  analysis}.
\newblock Griffin (London), 1971.

\bibitem{FJ:1981}
M.L. Fisher and R.~Jaikumar.
\newblock A generalized assignment heuristic for solving the vrp.
\newblock {\em Networks}, 11:109--124, 1981.

\bibitem{FR:1976}
B.~A. Foster and D.M. Ryan.
\newblock An integer programming approach to the vehicle scheduling problem.
\newblock {\em Operations Research}, 27:367–384, 1976.

\bibitem{Gaskell:1967}
T.~Gaskell.
\newblock Bases for vehicle fleet scheduling.
\newblock {\em Operational Research Quarterly}, 18:281--295, 1967.

\bibitem{GHL:1994}
M.~Gendreau, A.~Hertz, and G.~Laporte.
\newblock A tabu search heuristic for the vehicle routing problem.
\newblock {\em Management Science}, 40:1276--1290, 1994.

\bibitem{GLP:1999}
M~Gendreau, G~Laporte, and J~Potvin.
\newblock Metaheuristics for the vehicle routing problem.
\newblock Technical report, Les Cahiers du Gerad, 1998, revised 1999.

\bibitem{GM:1974}
B.E. Gillett and L.R. Miller.
\newblock A heuristic algorithm for the vehicle dispatch problem.
\newblock {\em Operations Research}, 22:340--349, 1974.

\bibitem{HK:1985}
M~Haimovich and A~H G~R Kan.
\newblock Bounds and heuristics for capacitated routing problems.
\newblock {\em Mathematics of Operations Research}, 10:527--542, 1985.

\bibitem{Hamilton:1856}
W.R. Hamilton.
\newblock Memorandum respecting a new system of roots of unity (the icosian
  calculus).
\newblock {\em Philosophical Magazine}, 12, 1856.

\bibitem{Holland:1975}
J.~H. Holland.
\newblock Adaptation in natural and artificial systems.
\newblock {\em The University of Michigan Press}, 1975.

\bibitem{Kar72}
R.~Karp.
\newblock Reducibility among combinatorial problems.
\newblock In R.~Miller and J.~Thatcher, editors, {\em Complexity of Computer
  Computations}, pages 85--103. Plenum Press, 1972.

\bibitem{Kirkman:1856}
T.P. Kirkman.
\newblock On the representation of polyhedra.
\newblock {\em Philosophical Transactions of the Royal Society of London Series
  A}, 146:413--418, 1856.

\bibitem{LMN:1986}
G.~Laporte, H.~Mercure, and Y.~Nobert.
\newblock An exact algorithm for the asymmetrical capacitated vehicle routing
  problem.
\newblock {\em Networks}, 16:33--46, 1986.

\bibitem{LANO:87}
G.~Laporte and Y.~Nobert.
\newblock {\em Surveys in Combinatorial Optimization}, chapter Exact algorithms
  for the vehicle routing problem.
\newblock 1987.

\bibitem{Laporte:1999}
G.~Laporte and F.~Semet.
\newblock Classical heuristics for the vehicle routing problem.
\newblock Technical report, Les Cahiers du Gerad, 1999.

\bibitem{LJDS:2009}
Chee~Peng Lim, Lakhmi~C. Jain, and Satchidananda Dehuri, editors.
\newblock {\em Innovations in Swarm Intelligence}, volume 248 of {\em Studies
  in Computational Intelligence}.
\newblock Springer, 2009.

\bibitem{Nagata:2007}
Y.~Nagata.
\newblock Edge assembly crossover for the capacitated vehicle routing problem.
\newblock {\em EvoCOP, LNCS}, 4446:142--153, 2007.

\bibitem{RTFNZ}
Road Transport~Forum NZ.
\newblock Road transport forum nz---transport facts.
\newblock Website.
\newblock \texttt{http://www.rtfnz.co.nz/cm-transport-facts.php}.

\bibitem{OSH:1987}
I.M Oliver, D.J. Smith, and J.R.C Holland.
\newblock A study of permutation crossover operators on the traveling salesman
  problem.
\newblock {\em Grefenstette---Proceedings of the Second International
  Conference on Genetic Algorithms and Their Applications}, pages 224--230,
  1987.

\bibitem{Or:1976}
I~Or.
\newblock {\em Traveling salesman-type combinatorial problems and their
  relation to the logistics of regional blood banking}.
\newblock PhD thesis, Evanston, IL: Northwestern University, 1976.

\bibitem{Osman:1993}
I.~H. Osman.
\newblock Metastrategy simulated annealing and tabu search algorithm for the
  vehicle routing problem.
\newblock {\em Annals of Operations Research}, 41:421--451, 1993.

\bibitem{Paessens:1988}
H.~Paessens.
\newblock The savings algorithm for the vehicle routing problem.
\newblock {\em European Journal of Operational Research}, 34:336--344, 1988.

\bibitem{PGKORZ:2005}
Pham, Ghanbarzadeh, Koc, S.~Otri, S.~Rahim, and M.~Zaidi.
\newblock The bees algorithm.
\newblock Technical report, 2005.

\bibitem{Potvin:2009}
Jean-Yves Potvin.
\newblock A review of bio-inspired algorithms for vehicle routing.
\newblock In {\em Bio-inspired Algorithms for the Vehicle Routing Problem},
  pages 1--34. 2009.

\bibitem{RZ:1968}
M.~R. Rao and S.~Zionts.
\newblock Allocation of transportation units to alternative trips—a column
  generation scheme with out-of-kilter subproblems.
\newblock {\em Operations Research}, 16:52--63, 1968.

\bibitem{RSD:2002}
M.~Reimann, M.~Stummer, and K.~Doerner.
\newblock A savings-based ant system for the vehicle routing problem.
\newblock {\em Proceedings of the Genetic and Evolutionary Computation
  Conference}, pages 1317--1325, 2002.

\bibitem{RBL:1996}
J~Renaud, F~F Boctor, and G~Laporte.
\newblock A fast composite heuristic for the symmetric traveling salesman
  problem.
\newblock {\em INFORMS Journal on Computing}, 8:134--143, 1996.

\bibitem{Robinson:1949}
J.~Robinson.
\newblock On the hamiltonian game (a traveling salesman problem).
\newblock {\em Research Memorandum RM-303}, 1949.

\bibitem{RDS:1990}
F.~Robuste, C.F. Daganzo, and R.~Souleyrette II.
\newblock Implementing vehicle routing models.
\newblock {\em Transportation Research}, 24B:263--286, 1990.

\bibitem{Ropke:2005}
Stefan Ropke.
\newblock {\em Heuristic and exact algorithms for vehicle routing problems}.
\newblock PhD thesis, Department of Computer Science, University of Copenhagen
  (DIKU), 2005.

\bibitem{Schrijver}
Alexander Schrijver.
\newblock On the history of combinatorial optimization (till 1960).
\newblock In {\em Operations Research and Management}. Elsevier, 2005.

\bibitem{Shaw:1998}
P.~Shaw.
\newblock Using constraint programming and local search methods to solve
  vehicle routing problems.
\newblock pages 417--431. 1998.

\bibitem{Solomon:1987}
M.~M. Solomon.
\newblock Algorithms for the vehicle routing and scheduling problems with time
  window constraints.
\newblock {\em Oper. Res.}, 35(2):254--265, 1987.

\bibitem{Taillard:1993}
E.~D. Taillard.
\newblock Parallel iterative search methods for vehicle routing problems.
\newblock {\em Networks}, 23:661--673, 1993.

\bibitem{TNJ:1991}
S.R. Thangiah, K.E. Nygard, and P.L. Juell.
\newblock Gideon: A genetic algorithm system for vehicle routing with time
  windows.
\newblock {\em Proceedings of 7th IEEE Conference on Artificial Intelligence
  Applications}, pages 322--328, 1991.

\bibitem{GHL:1998}
P.~Toth and D.~Vigo.
\newblock The granular tabu search (and its application to the vehicle routing
  problem).
\newblock Technical report, 1998.

\bibitem{TV2001}
Paolo Toth and Daniele Vigo, editors.
\newblock {\em The vehicle routing problem}.
\newblock Society for Industrial and Applied Mathematics, Philadelphia, PA,
  USA, 2001.

\bibitem{Willard:1989}
J.~A.~G. Willard.
\newblock Vehicle routing using r-optimal tabu search.
\newblock Master's thesis, The Management School, Imperial College, London,
  1989.

\bibitem{WH:1972}
A.~Wren and A.~Holliday.
\newblock Computer scheduling of vehicles from one or more depots to a number
  of delivery points.
\newblock {\em Operations Research Quarterly}, pages 333--344, 1972.

\bibitem{Yellow:1970}
P.~Yellow.
\newblock A computational modification to the savings method of vehicle
  scheduling.
\newblock {\em Operational Research Quarterly}, 21:281--283, 1970.

\end{thebibliography}


\end{document}